\documentclass[lettersize,journal]{IEEEtran}
\usepackage{amsmath,amsfonts}
\usepackage{array}
\usepackage[caption=false,font=normalsize,labelfont=sf,textfont=sf]{subfig}
\usepackage{textcomp}
\usepackage{stfloats}
\usepackage{url}
\usepackage{verbatim}
\usepackage{graphicx}
\usepackage{cite}
\usepackage{amsthm}
\usepackage{amssymb}

\usepackage{multicol}

\usepackage{changepage}

\usepackage{makecell}

\usepackage{diagbox}

\usepackage{tabularx}

\usepackage{bbm}

\usepackage{mathrsfs} 
\usepackage{subfig}

\usepackage[hidelinks]{hyperref}

\DeclareMathOperator*{\argmin}{arg\,min}
\DeclareMathOperator*{\argmax}{arg\,max}
\usepackage{algorithm}
\usepackage{algorithmicx}
\usepackage{algpseudocode}

\begin{document}

\title{Stealthy Poisoning Attacks Bypass Defenses in Regression Settings}

\author{Javier Carnerero-Cano, Luis Mu\~noz-Gonz\'alez, Phillippa Spencer, and Emil C. Lupu
        \thanks{J. Carnerero-Cano is with IBM Research Europe, Portal, First Floor Trinity Business School, Trinity College Dublin, Dublin, D02 F6N2, Ireland. Work done while with Imperial College London, South Kensington Campus,  London, SW7 2AZ, United Kingdom. E-mail: javier.cano@ibm.com.}
        \thanks{L. Mu\~noz-Gonz\'alez is with Universidad de Alcalá de Henares, Escuela Politécnica Superior, Alcalá de Henares, Madrid, 28805, Spain. E-mail: luis.munozg@uah.es. His work is supported by the Spanish Ministry of Science and Innovation under the ATRAE Program (ATR2024-154987).}
                \thanks{P. Spencer is with BAE Systems Air, United Kingdom. Work done while with the Defence Science and Technology Laboratory (DSTL), Porton Down, Salisbury, United Kingdom.}
\thanks{E. C. Lupu is with Imperial College London, South Kensington Campus,  London, SW7 2AZ, United Kingdom. E-mail: e.c.lupu@imperial.ac.uk.}\protect\\
}

\markboth{Journal of \LaTeX\ Class Files}%
{Shell \MakeLowercase{\textit{et al.}}: A Sample Article Using IEEEtran.cls for IEEE Journals}


\maketitle

\begin{abstract}
 Regression models are widely used in industrial processes, engineering, and in natural and physical sciences, yet their robustness to poisoning has received less attention. When it has, studies often assume unrealistic threat models and are thus less useful in practice. In this paper, we
 propose a novel optimal \textit{stealthy} attack formulation that considers different degrees of detectability and show that it bypasses state-of-the-art defenses.
 We further propose a new methodology based on normalization of objectives to evaluate different trade-offs between effectiveness and detectability. Finally, we develop a novel defense (\textit{BayesClean}) against stealthy attacks. BayesClean improves on previous defenses when attacks are stealthy and the number of poisoning points is significant.
\end{abstract}


\section{Introduction}
\IEEEPARstart{R}{egression}  is widely used in mission-critical systems including in the monitoring of aircraft engines \cite{chati2017gaussian}, for the development of pharmaceuticals \cite{ekins2019exploiting},
in the management of hedge funds \cite{meligkotsidou2009quantile}, and in predictive maintenance \cite{wen2022recent} and quality control \cite{sun2017improved} for industrial processes. In such contexts, the training data vulnerable to poisoning as it is often collected from partially trusted distributed supply chains or from IoT devices that are easy to compromise.

Despite this, the robustness of regression models to poisoning attacks has not been widely studied. Most studies on this topic \cite{mei2015using, zhang2018training, jagielski2018manipulating, muller2020data, wen2021great, weerasinghe2021closing, li2021optimal}, assume the attacker's objective is solely to maximize the damage, without considering attacks may aim to be \textit{stealthy}, i.e., less detectable. In this paper, we show that the defenses proposed in these studies fail to mitigate stealthy attacks.
We argue that in practical scenarios, attackers will seek to maximize the attack effectiveness while remaining undetected, because if poisoning is detected, the training set may be discarded thwarting the attack entirely. Although \textit{stealthy} attacks may be more costly and less effective compared to unconstrained attacks, they can produce a significant more damage in practice as they are more difficult to detect.

We consider the following research questions: \textbf{RQ1}: Can we craft attacks capable of different trade-offs between effectiveness and detectability that bypass current defenses? \textbf{RQ2}: Can we design more effective defenses capable of mitigating stealthy attacks?

To answer the first question, we propose a novel formulation of poisoning attacks as a parameterizable combination of attack objectives: (i) maximizing the damage to the model and (ii) minimizing the risk of being detected. We model the attacker's problem as a multiobjective bilevel optimization problem and show how it can be solved. This formulation is more general than \textit{adaptive attacks}, which target specific weaknesses of the defense, and allows us to construct attacks that simultaneously evade different types of defenses. For this, we exploit attack transferability,
and show that the appropriate modeling of detectability constraints, targeting a standard model, is enough to bypass the defenses. Furthermore, we show that in some cases defenses are not only unable to mitigate the attack, but also damage the model's performance compared to an unprotected model. In other words, the use of these defenses can be detrimental.

We focus on Linear Regression (LR), as most existing literature focuses on this setting. LR is also widely used in practice, so it is important to defend these models \cite{marill2004advanced, jagielski2018manipulating, montgomery2021introduction, weerasinghe2021closing}. The analysis of the robustness of linear models also allows us to gain more general insights about attack effectiveness and limitations of existing defense strategies. Beyond LR, we also demonstrate these insights by also evaluating the performance and scalability of our attack in nonlinear models like Deep Neural Networks (DNNs).

Our results show that noise and uncertainty about predictions play a pivotal role in the success of stealthy poisoning attacks. Thus, based on this observation we propose \emph{BayesClean} a novel defense capable of overcoming some of the limitations of existing defenses to mitigate stealthy attacks and, consequently, answering RQ2. We show that \emph{BayesClean} is capable of reducing the impact of stealthy poisoning attacks, by rejecting points based on their location within the predictive variance of the model. This principle is more promising for detecting and rejecting suspicious points compared to previous approaches and can form the basis for the development of further defenses.

In summary, the contributions of this paper are as follows:
\begin{itemize}
    \item We propose a novel stealthy poisoning attack formulation against regression models via multiobjective bilevel optimization: One objective accounts for attack effectiveness, the other for the detectability of poisoned points. For this, we introduce a novel detectability-risk function and a novel approach for solving the optimization problem.

    \item We show that state-of-the-art defenses fail to mitigate the stealthy attacks we propose, even when they are not adaptive. We evaluate the different defenses proposed so far in the literature on real-world datasets, in both LR models and DNNs, and identify the factors that influence their effectiveness. To our knowledge, we are the first to analyze optimal indiscriminate poisoning attacks targeting DNNs in regression settings.

    \item We propose \emph{BayesClean}, a novel defense based on uncertainty and Bayesian models, that can better detect and reject malicious datapoints even when they are stealthy. We show that \emph{BayesClean} improves upon state-of-the-art defenses against stealthy poisoning attacks. This also evidences the importance of the model uncertainty for mitigating poisoning attacks in regression.

\end{itemize}

\section{Related Work}
Most existing poisoning attacks against regression aim to maximize the model's Mean Squared Error (MSE)
\cite{mei2015using, zhang2018training, jagielski2018manipulating, muller2020data, wen2021great, weerasinghe2021closing}.
 Mei and Zhu \cite{mei2015using} poison convex learning algorithms, including LR, based on the concept of machine teaching \cite{zhu2013machine,zhu2018overview}. Jagielski~et~al.~\cite{jagielski2018manipulating} propose Perfect-Knowledge (PK) and Limited-Knowledge (LK) poisoning attacks against LR aiming to increase the MSE of the model at test time. The PK attack is based on a bilevel optimization formulation solved by leveraging its Karush-Kuhn-Tucker (KKT) conditions, similar to Xiao~et~al.~\cite{xiao2015feature}. Nevertheless, the computation of the hypergradients of the attacker (gradients of the attacker's objective function) is expensive due the inversion of the Hessian w.r.t. the model's parameters.
 They also propose an LK attack, called \emph{StatP}, that
samples points from a multivariate Gaussian distribution
estimated from the clean training data \cite{jagielski2018manipulating}. StatP rounds the feature variables to the bounds of their feasible  domain, queries the model, and rounds the target variable to the opposite corner.
However, these attacks can be detected by defenses that reject extreme values \cite{jagielski2018manipulating, wen2021great}. Weerasinghe~et~al.~\cite{weerasinghe2021closing} incrementally extend the attack formulation of Jagielski~et~al.~\cite{jagielski2018manipulating} to LR models that use nonlinear basis functions. Wen~et~al.~\cite{wen2021great} also slightly modify the optimal poisoning attack in \cite{jagielski2018manipulating} by including the influence of previous poisoning points in the attacker's objective. M\"uller~et~al.~\cite{muller2020data} propose an LK label-manipulation attack with a surrogate training set, flipping the labels to one of two possible thresholds.

However, an attacker's objective in poisoning a regression model may also differ. For example,
attacks targeting specific parameters of an LR model have also been considered in \cite{li2021optimal}.

In terms of defenses, algorithms for robust regression that are not strongly affected by the presence of stochastic noise and outliers are available in robust statistics \cite{huber1964robust, fischler1981random}. However, these methods fail in high dimensions and struggle to identify carefully crafted poisoning samples \cite{xu2012outlier}. Liu~et~al.~\cite{liu2017robust} propose robust LR algorithms when the feature matrix has low rank and can be projected to a lower dimensional space. Diakonikolas~et~al.~\cite{diakonikolas2019sever} propose SEVER, a robust meta-algorithm that can detect poisoning samples through outlier scores based on the singular value decomposition of the gradients. Weerasinghe~et~al.~\cite{weerasinghe2021closing} propose BIG-LID, an LK defense for nonlinear regression learning based on the concept of Local Intrinsic Dimensionality (LID)\footnote{The LID value of a data point identifies the minimum number of latent variables required to represent that data point and recent evidence suggests that LID is an indicator of the degree of being an outlier \cite{houle2018correlation, weerasinghe2022local}.} \cite{houle2017local} to identify and reduce the impact of poisoning samples. All these methods have provable robustness guarantees, but the assumptions on which they rely are not usually satisfied in practical settings where attackers also consider attack detectability.

Other relevant defenses against poisoning attacks in regression include TRIM \cite{jagielski2018manipulating} and Proda \cite{wen2021great}. TRIM \cite{jagielski2018manipulating}

iteratively finds a subset of the training data that minimizes the model's loss, and uses this subset to train the
model. Similar to aggregation methods, Proda \cite{wen2021great} filters poisoning points by bootstrapping equal-sized groups of training points. Each bootstrap set is independently subjected to LR, resulting in $\beta$ trained models. Then, for each trained model, Proda selects the subset of the training set with the lowest errors on the trained model, which is subjected again to LR. Among these $\beta$ trained models, Proda selects the model with the lowest MSE.

Finally, Debugging Using Trusted Items (DUTI) \cite{zhang2018training} identifies samples that have the biggest influence on the validation error evaluated on a small trusted clean set.

It is important to note that most of these defenses assume that they know the fraction of poisoning samples in the training dataset. This is unrealistic in practical applications. Furthermore, deviations on the estimation of the fraction of poisoning points can  negatively impact the model's performance for both poisoned and clean models. In contrast, our \emph{BayesClean} defense does not require estimating the fraction of poisoning points to reject, as we reject points according to the predictive variance of a Bayesian LR model.

\section{Threat Model}

In regression, given an input space, $\mathcal{X}\subseteq \mathbb{R}^m$, and a continuous label space, $\mathcal{Y}\subseteq \mathbb{R}$, the learner aims to estimate the mapping $f: \mathcal{X} \rightarrow \mathcal{Y}$. This mapping is typically approximated with a model $\mathcal{M}$ that depends on a set of parameters or hyperparameters ${\bf w}\in \mathbb{R}^d$ learnt using a training dataset $\mathcal{D}_\mathrm{tr}= \{({\bf x}_i , y_i)\}^{n_\mathrm{tr}}_{i=1}=({\bf X}_{\mathrm{tr}},{\bf y}_{\mathrm{tr}})$, where $n_\mathrm{tr}$ is the number of training samples. Typically, training points are assumed to be independent, identically distributed (IID), and sampled from the underlying distribution $p(\mathcal{X}, \mathcal{Y})$.
In LR, the model's output is a linear function $f({\bf x}, {\bf w}, b) = {\bf w}^{\textsf{T}} {\bf x} + b$, where $\bf{w}$ and the bias term $b$ are the model parameters. The output $f({\bf x}, {\bf w}, b)$ aims to predict the value of $y$ for input $\bf{x}$. For the sake of clarity, both for linear and nonlinear regression models, we use $f({\bf x}, {\bf w})$ to define the model's output, assuming the bias terms are already contained in the vector of parameters ${\bf w}$.

Given that in regression the target is a continuous variable, the model parameters are typically learned by minimizing the MSE evaluated on the training data points. Then, the loss function, $\mathcal{L}(\mathcal{D}_\mathrm{tr}, {\bf w})$ can be written as:
\begin{equation}
    \mathcal{L}(\mathcal{D}_\mathrm{tr}, {\bf w}) = \frac{1}{2 \ n_\mathrm{tr}} \sum_{i = 1}^{n_\mathrm{tr}} (f({\bf x}_i, {\bf w}) - y_i)^2 + \lambda \Omega({\bf w}),
\end{equation}
where the first term represents the MSE evaluated on the training set and $\Omega({\bf w})$ is the regularization term controlled by the hyperparameter $\lambda$ commonly used to prevent overfitting. $\lambda$ can be adjusted by using cross validation or by minimizing the MSE on a separate validation dataset. In this paper we use $L_2$ regularization, i.e., \emph{ridge regression}.

Although existing work on the robustness of regression has mostly considered LR models, the threat model and attack formulation proposed here are also applicable to more general (nonlinear) models, such as DNN architectures. Similarly to other works in data poisoning, we characterize adversaries according to their capabilities, knowledge, and goals.

\subsection{Attacker's Capabilities}
We assume an adversary able to compromise a fraction of the training set to achieve certain malicious goals \cite{barreno2006can,barreno2010security} by injecting $n_p$ poisoning points in the training set, so that the total number of training points is $n'_\mathrm{tr} = n_\mathrm{tr} + n_p$, where $n_\mathrm{tr}$ is the number of clean (non-compromised) points. The fraction of poisoning points injected is then $n_p / (n_\mathrm{tr} + n_p)$.

Depending on the attacker's capabilities to manipulate the training data, we distinguish three possible scenarios: (i) \emph{Label manipulation attacks}, where the attacker is only able to modify the labels of the data, which, in this case are continuous. (ii) \emph{Feature manipulation attacks}, where the attacker can manipulate the features of the input but not the labels. (iii) \emph{Data manipulation attacks}, where the attacker can manipulate both features and labels. Similarly to other studies, we consider data manipulation attacks, as they provide a more general view of the problem.

However, in contrast with related work, we consider scenarios with different poisoning regimes, including cases where attackers can poison a significant fraction of the training points. We argue that such scenarios are realistic because training data is often acquired from from sensors, IoT or OT equipment that can be compromised. Due to the homogeneity of such equipment -- typically only a few suppliers are used -- many devices have the same vulnerabilities and can be compromised at negligible additional cost for the attacker. The attacker therefore can trade-off between cost, detectability and attack success. If the attack is detected, the training set may be discarded. So, to ensure success, the attack must be ``stealthy'' i.e. difficult to detect. So, it is often preferable to poison more points with smaller perturbations to achieve the same effect for a small additional cost.

\subsection{Attacker's Knowledge}

We consider an attacker aware of the training data, the feature set, the model, and the objective function---a worst case scenario. However, we do not assume knowledge of the defense used by the victim. Thus, we are in a Limited Knowledge (LK) threat model where the attacker targets an unprotected regression model, and we evaluate the attack effectiveness against models that use different defenses. We chose this setting because we aim to use a more general formulation of the poisoning attack that takes into account the risk of being detected. In contrast, defensive algorithms proposed in the literature differ in their assumptions or assume adaptive attacks where the attacker has knowledge of the specific defense deployed. For example, DUTI, requires the defender to have access to a clean validation set, whereas other approaches, like TRIM, do not. Given the heterogeneity of these methods, rather than crafting different adaptive attacks targeting specific defenses, we use the same attack algorithm to show that such defenses fail to detect and mitigate \textit{stealthy} attacks that seek to evade detection. While stealthy attacks may be sub-optimal compared to adaptive attacks in Perfect Knowledge (PK) settings, they nevertheless allow attackers to reach their goal, especially in LK scenarios where attackers are not aware of the target model and the defense used.
Our results show that our general formulation allows us to bypass existing defenses, and, shockingly, that in some cases, defenses lead to worse performance than an unprotected model.

\subsection{Attacker's Goal and Strategy}
In data poisoning attacks the adversary aims to manipulate the behavior of the target model by injecting malicious points in the training dataset, typically to increase the error on the model's predictions. Attacks can be \emph{indiscriminate} (poisoning availability attack) when they affect predictions for many samples or \emph{targeted} (poisoning integrity attacks) when they manipulate predictions for a target set of data points \cite{barreno2006can,huang2011adversarial}. Depending on the errors to be produced, we differentiate between \emph{error-generic attacks}, where the attacker does not care about the errors produced by the attack, and \emph{error-specific attacks}, where the attacker aims to produce a specific type of error in the predictions \cite{munoz2017towards, munoz2019challenges}, e.g., to overestimate the target variable. Consistently with previous works in the literature on poisoning regression, we study error-generc attacks.

To achieve such an attack goal, a poisoning attack can adopt different strategies that can be characterized by an \textit{effectiveness-detectability} trade-off. Highly effective attacks seek to minimize the number of required poisoning points \cite{biggio2012poisoning, mei2015using, munoz2017towards, jagielski2018manipulating, carnerero2021regularization, carnerero2024hyperparameter}. However, to achieve the attack objective, poisoning points can differ significantly from the clean points, which makes them highly detectable \cite{paudice2018detection}. This makes it easier to detect the attack and can thwart it entirely if the defender discards the poisoned training set. Whilst agressive attack are often encountered when it is too late for the defender to intervene, e.g., Denial-of-Service, data exfiltration or damaging a cyber-physical system \cite{narula2025exploring}, when poisoning  the training set of an ML algorithm, the attack needs to remain undetected for the damage to be caused. To remain undetected, poisoning points need to be constrained by a set of \textit{detectability constraints}, reflecting the attacker's assumptions about the defense and its tolerances. Considering detectability constraints invariably blunts the effectiveness of each poisoning point and requires more poisoning points to achieve the attack objectives, which remain nevertheless achievable.

In the research literature on poisoning attacks, detectability has been considered in different attacks for classification tasks e.g. clean-label poisoning attacks \cite{shafahi2018poison,geiping2021witches} or Generative Adversarial Nets \cite{munoz2019poisoning}, and in some model poisoning attacks for federated learning \cite{bhagoji2019analyzing,baruch2019little}. However, to our best knowledge, detectability constraints have not been adequately modeled for poisoning attacks against regression algorithms.

We do not make assumptions on the detectability constraints an attacker may consider, but propose a formulation that allows us to investigate different \textit{effectiveness-detectability} trade-offs by balancing between the effectiveness and the detectability objectives. We characterize the detectability constraints by assuming that an initial set of poisoning points is given, and that the poisoning points are modified according to a space of possible modifications, e.g., by constraining the norm of the perturbation of each poisoning point. This space of possible modifications can then be included in the attack formulation we introduce in the next section. Note that detectability constraints in regression settings generalize to those for classification, since the detectability constraints for the labels are also continuous.

\section{Poisoning Attacks with Detectability Constraints}
\label{section:attacks_soft}
Under PK settings, we formulate the optimal poisoning attack strategy, i.e., the one that maximizes the impact on the target model, as a bilevel optimization problem:

\begin{equation}
\begin{aligned}
\max_{\mathcal{D}_\mathrm{p}' \in \Phi(\mathcal{D}_\mathrm{p})}  \quad & \mathcal{A}(\mathcal{D}_\mathrm{target},    {\bf w}^\star)  \\
\mathrm{s.t.} \quad &  {\bf w}^\star \in \argmin_{{\bf w} \in \mathcal{W}} \mathcal{L}(\mathcal{D}_\mathrm{tr} \cup \mathcal{D}_\mathrm{p}', {\bf w}),\\
\end{aligned}
\label{eq:opt_attack}
\end{equation} where the outer problem represents the attacker's objective,
and the inner problem is the defender's problem, i.e., training the model with the poisoned training set. We assume that the defender trains on all the data points $\mathcal{D}_\mathrm{p}$ that are in the feasible set $\Phi(\mathcal{D}_\mathrm{p})$, i.e., they do not discard points as they do not have prior knowledge of the attack.

This formulation encompasses both \textit{indiscriminate attacks}, where  $\mathcal{D}_\mathrm{target}=\mathcal{D}_\mathrm{val}$, and the validation set is representative of the underlying clean data distribution, and \textit{targeted attacks}, where $\mathcal{D}_\mathrm{target}$ represents a specific subset of target data points.\footnote{Note that targeted attacks against LR also impact other points that are not in the attacker's target.} Generally, $\mathcal{A}$ represents the function for the attacker objective. As we aim to maximize the predictions' error indiscriminately, we use the MSE, i.e., the loss typically minimized by most regression models $\mathcal{A}(\mathcal{D}_\mathrm{target}, {\bf w}^\star)=\mathcal{L}(\mathcal{D}_\mathrm{val}, {\bf w}^\star)$ to model the attacker's objective.

Given a attacker's budget of poisoning points, $n_\mathrm{p}$, each poisoning point can be optimized by solving Eq. (\ref{eq:opt_attack}) incrementally (point by point), in mini-batches, or for all poisoning points at the same time. In the two former cases, $\mathcal{D}_\mathrm{tr}$ contains the previous poisoning points, which influences the computation of subsequent points. This formulation is often encountered in related studies on poisoning attacks both for regression and classification \cite{biggio2012poisoning, xiao2015feature, mei2015using, munoz2017towards, jagielski2018manipulating, muller2020data, wen2021great, weerasinghe2021closing, carnerero2021regularization, carnerero2024hyperparameter}.

\subsection{Optimal Attack Formulation with Soft Detectability Constraints}
\label{subsec:attack_with_detectability}

In addition to maximizing damage, an attacker also seeks to remain undetected. So, a second, but equally important attacker objective is to reduce the \textit{detectability risk}. This may reduce the effectiveness of the attack, but is necessary to avoid completely jeopardizing the attack. The attacker's perception of the risk of being detected constrains the attack effectiveness and restricts the manipulation of the features and labels of the poisoning points.
Thus, the detectability-risk objective is to decrease
the poisoned points detected as malicious and, for some aggressive defenses, the clean points detected as benign.

There is no general agreement on how to define the detectability risk (or detectability constraints). \textit{Hard} constraints,\footnote{In an optimization problem, a hard constraint is a constraint that must be satisfied by any feasible solution to the model. On the other hand, a soft constraint can be violated, but violating the constraint incurs a penalty in the objective function (often, the greater the amount by which the constraint is violated, the greater the penalty) \cite{meseguer2006soft}.} typically considered in adversarial ML literature \cite{suciu2018does, demontis2019adversarial,  huang2020metapoison, huang2021unlearnable, geiping2021witches, fowl2021adversarial, koh2022stronger, cina2023wild}, must be satisfied by any feasible solution. In practice, the attacker's willingness to reduce attack effectiveness to avoid detection depends on the attacker's risk appetite. In particular,
the adversary may not be aware of the defense algorithm. Even if the defense algorithm is known and has a concrete threshold for removing points \cite{jagielski2018manipulating, diakonikolas2019sever}, the rejection region can be very complex.

 Some formulations that consider hard constraints defined by a particular defense, e.g., \cite{koh2022stronger}, are intractable and difficult to optimize in practice. So, we adopt a \textit{soft} (parameterizable) formulation of the \emph{detectability risk},  $\mathcal{R}$,
 which also allows to model different trade-offs between effectiveness and detectability that are easier to interpret.

We thus revisit the bilevel formulation of the optimal attack to include the detectability risk. We define the outer objective as a \emph{multiobjective optimization problem} where one of the objectives is the MSE of the model (effectiveness of the attack), and the other is the detectability risk $\mathcal{R}$:
\begin{equation}
\begin{aligned}
\max_{\mathcal{D}_\mathrm{p}' \in \Phi(\mathcal{D}_\mathrm{p})}  \mathcal{L}(\mathcal{D}_\mathrm{val}, &   {\bf w}^\star),  \min_{\mathcal{D}_\mathrm{p}' \in \Phi(\mathcal{D}_\mathrm{p})}    \mathcal{R} \\
\mathrm{s.t.} \quad &  {\bf w}^\star \in \argmin_{{\bf w} \in \mathcal{W}} \mathcal{L}(\mathcal{D}_\mathrm{tr} \cup \mathcal{D}_\mathrm{p}', {\bf w}).
\end{aligned}
\end{equation}

Effectiveness and detectability are competing objectives. So, a solution to the multiobjective problem must quantify their trade-off. A straightforward way to achieve this is by using scalarization \cite{marler2004survey}, i.e.,

varying the ``importance'' of each objective by a scalar $\alpha \in [0, 1]$:
\begin{equation}
\begin{aligned}
\max_{\mathcal{D}_\mathrm{p}' \in \Phi(\mathcal{D}_\mathrm{p})}  \quad & \mathcal{A}_d (\mathcal{D}_\mathrm{p}') = \alpha \mathcal{L}(\mathcal{D}_\mathrm{val}, {\bf w}^\star) - (1 - \alpha)\mathcal{R} \\
\mathrm{s.t.} \quad &  {\bf w}^\star \in \argmin_{{\bf w} \in \mathcal{W}} \mathcal{L}(\mathcal{D}_\mathrm{tr} \cup \mathcal{D}_\mathrm{p}', {\bf w}),\\
 \quad & \alpha \in [0, 1].
 \label{eqAttacker_}
\end{aligned}
\end{equation}

The hyperparameter $\alpha$ can be set by the attacker based on its risk appetite and a parametric sweep allows us to investigate different effectiveness detectability trade-offs. Note that, when combining objectives in this way, the correlation between the objective functions is known to affect the search space structure and algorithm performance. To enable a smooth trade-off, the outer objective $\mathcal{A}_d$ should admit multiple local optima rather than a single one.

\subsubsection{Modeling the Detectability-Risk Function}
\label{subsec:detectability_choice}

\begin{figure}[!t]
\centering
{\includegraphics[width=1.6in]{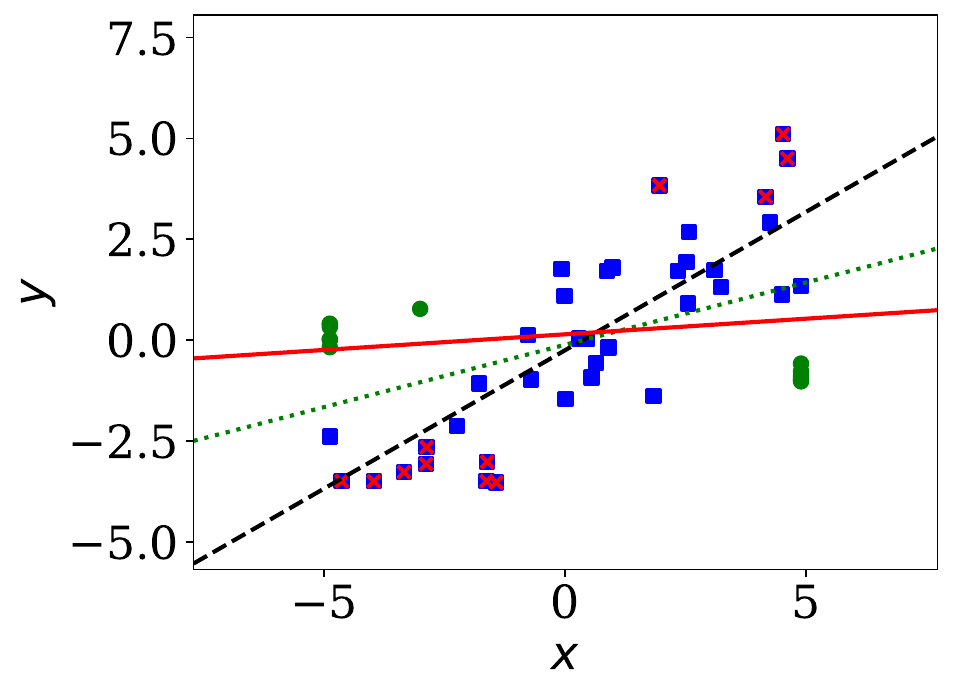}
\vspace{-0.4cm}
\label{fig:trim_a}}

\caption{Synthetic example showing the effect of constraining poisoning points against TRIM \cite{jagielski2018manipulating}. The blue points represent the clean points, the green points are the poisoning points, the red crosses are the points rejected by TRIM, the black dashed line is the regression line learned using the original clean data, the green dotted line is the regression line learned with the complete poisoned training data, and the red solid line is the regression line learned after applying TRIM. Because the poisoning points are close to the clean points, TRIM fails to detect them.\vspace{-0.3cm}}
\label{fig:trim}
\end{figure}

To evaluate the detectability risk, i.e., the attacker's perception of the risk of being detected, different formulations are possible, as defenders can aim to detect poisoning points in two different ways:

\begin{itemize}
    \item \textbf{Data distribution-based detectability}: poisoned points can be detected based on an an estimation of the underlying data distribution, regardless of the target model. For instance, detection can based on distance functions between points: $\mathcal{R}(\mathcal{D}_\mathrm{p}', \mathcal{D}_\mathrm{tr}, \mathcal{D}_\mathrm{val})$.

    \item \textbf{Model-based detectability}: poisoned points can also be detected based on the effect they have on the target model, for example, their impact on the loss function. In this case, the detectability risk, $\mathcal{R}(\mathcal{D}_\mathrm{p}', {\bf w}^\star, {\bf w}^\star_\mathrm{cl})$, is a function of the feasible poisoning points, $\mathcal{D}_\mathrm{p}'$, the poisoned parameters, ${\bf w}^\star$,  and the clean parameters, ${\bf w}_\mathrm{cl}^\star \in \argmin_{{\bf w} \in \mathcal{W}} \mathcal{L}(\mathcal{D}_\mathrm{tr}, {\bf w})$.
\end{itemize}

We choose the latter approach, noting that state-of-the-art defenses reject points whose loss or gradients' norm is large, e.g., \cite{jagielski2018manipulating, diakonikolas2019sever, wen2021great}. Such defenses assume poisoned points are outliers and fail when this is not the case. For example, we can see in Fig.~\ref{fig:trim}\footnote{Further experimental settings can be found in Appx.~\ref{subsec:expset2}.} that TRIM fails even when simply bounding the maximum values of the poisoned points. It rejects several genuine data points but none of the poisoned ones. The deviation of the model learned with TRIM w.r.t. the model trained with clean data is notably higher than the deviation of the unprotected model trained on the poisoned data. In other words, TRIM has a negative effect and can be worse than not using any defense.

\subsubsection{Design of the Detectability-Risk Function}
\label{subsec:practical}

For practical purposes, we craft poisoning points in mini-batches and optimize them incrementally. There are $n_{B_\mathrm{p}}$ poisoning points optimized in batches of $n_{B_\mathrm{p}} = \lceil {n_\mathrm{p}/B_\mathrm{p}}\rceil $ points. Once a batch is optimized, it is fixed in the training set and the next batch is optimized. We assume the defender has access to clean parameters and/or clean data when setting its detection (e.g., thresholds for anomaly detection are normally set on clean data), so the detectability risk does not change with the increasing number of poisoning points.

It is possible to design a detectability-risk function that minimizes the loss function on the poisoning points, i.e., $\mathcal{R}(\mathcal{D}_\mathrm{p}', {\bf w}^\star_\mathrm{cl})=\mathcal{L}(\mathcal{D}_\mathrm{p}', {\bf w}^\star_\mathrm{cl})$. However, this function and the attacker's goal function, $\mathcal{L}(\mathcal{D}_\mathrm{val}, {\bf w}^\star)$, are redundant, as both favor points close to the regression hyperplane.
So, we define the detectability risk as the product of two loss functions that penalize points near the bounds of a ``confidence interval'':
\begin{equation}
\begin{aligned}
 \mathcal{R}(\mathcal{D}_\mathrm{p}', {\bf w}^\star_\mathrm{cl}) = \mathcal{L}(\mathcal{D}_\mathrm{p}', {\bf w}^\star_{\mathrm{cl}_+})\mathcal{L}(\mathcal{D}_\mathrm{p}', {\bf w}^\star_{\mathrm{cl}_-}).
\end{aligned}
\label{eq:detecfunc}
\end{equation}

where ${\bf w}^\star_{\mathrm{cl}_+}$ and ${\bf w}^\star_{\mathrm{cl}_-}$ are all equal to the clean parameters ${\bf w}^\star_{\mathrm{cl}}$ except for their bias terms: $b_{\mathrm{cl}_+}^\star = b_\mathrm{cl}^\star + \sigma$ and $b_{\mathrm{cl}_-}^\star = b_\mathrm{cl}^\star - \sigma$; $\sigma$ is the residual standard deviation/error):
 $\sigma = \left({{\sum_{i=1}^{n_\mathrm{tr}}\left(y_{\mathrm{tr}_i}-f({\bf x}_{\mathrm{tr}_i})\right)^2}/{df}}\right)^{1/2}$, and $df$ are the degrees of freedom of the residuals, which in the case of LR are: $n_\mathrm{tr} - p - 1$.

 Defining the risk function (\ref{eq:detecfunc}) as a product of two functions, allows us to find non-trivial local maxima in the attacker’s objective (not in the corners of the attacker’s feasible domain or on the regression hyperplane) and a smoother transition between the two objectives.
By sweeping values of $\alpha$ in (\ref{eqAttacker_}) between $0$ to $1$, we obtain poisoning points that shift smoothly from inliers to outliers, balancing the two goals proposed: attack success and stealth. However, in practice, the two objectives can be of different orders of magnitude, which leads to one objective dominating the other. So we need to define a normalization method to balance the two objectives.

\subsubsection{Normalization of Objective Functions}

\label{subsubsec:norm_obj}

As the effectiveness and the detectability objective functions used in Eq.~(\ref{eqAttacker_}) can have values of different scale, normalization of the objectives is needed. This also provides interpretability for $\alpha$, i.e., the stealth factor of the attack.

\begin{figure}[!t]
\centering
\subfloat[]{\includegraphics[width=1.7in]{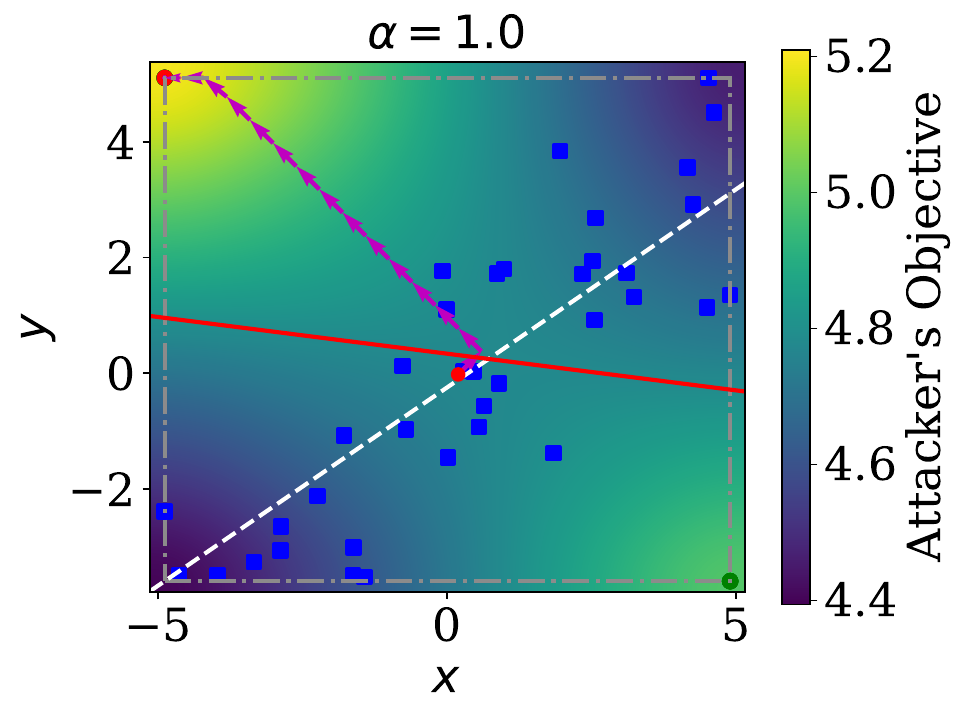}
\label{fig:no_norm_a}}
\subfloat[]{\includegraphics[width=1.7in]{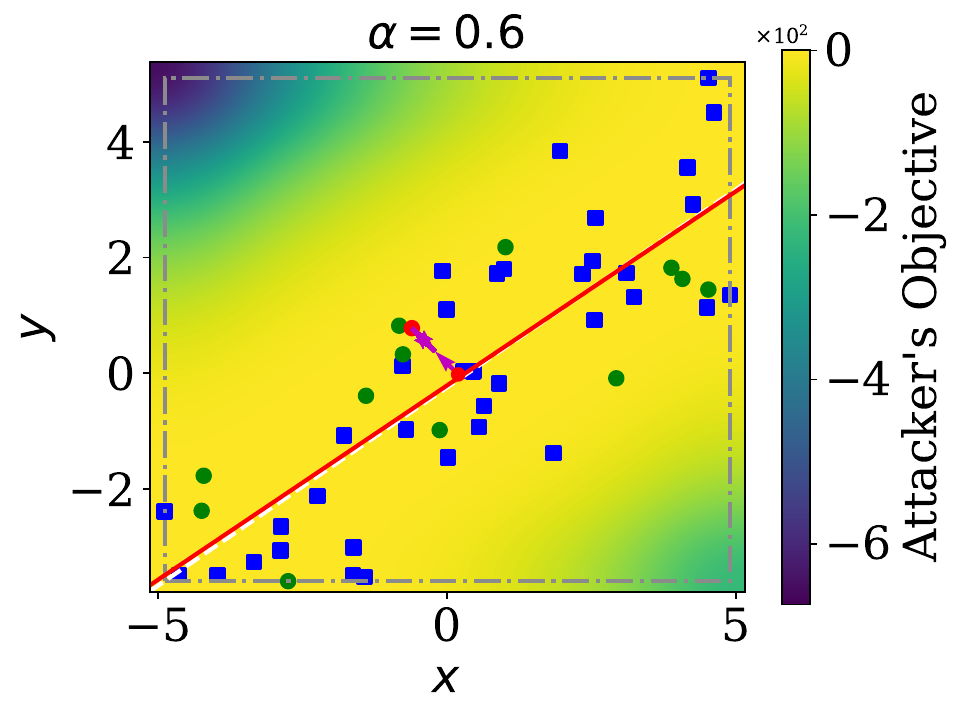}
\label{fig:no_norm_b}}
\\
\vspace{-0.4cm}
\subfloat[]{\includegraphics[width=1.7in]{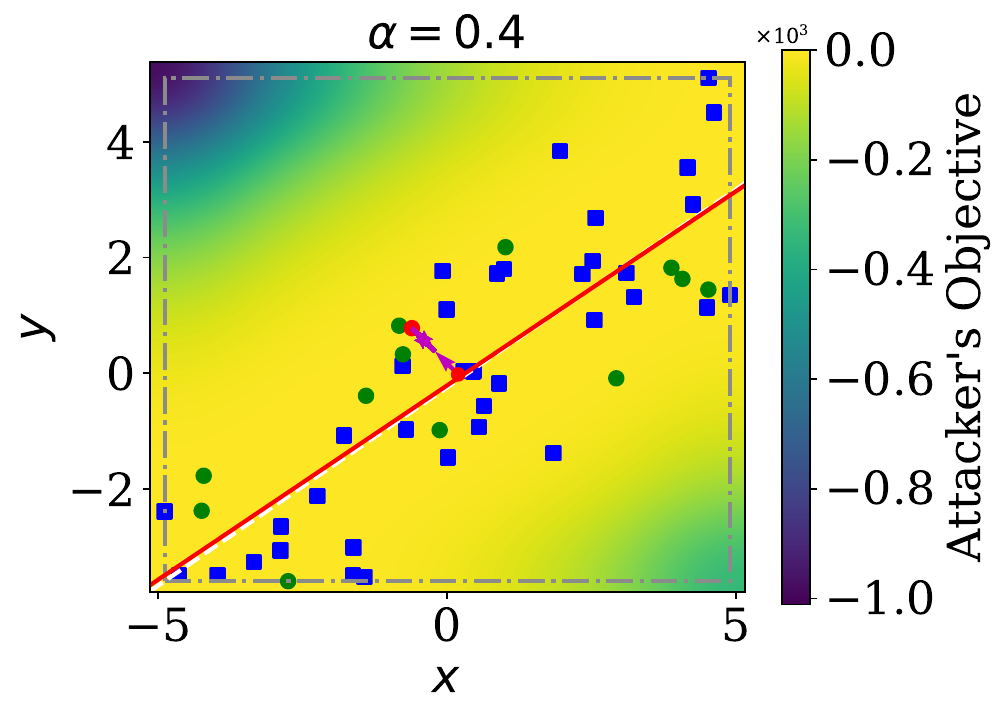}
\label{fig:no_norm_c}}
\subfloat[]{\includegraphics[width=1.7in]{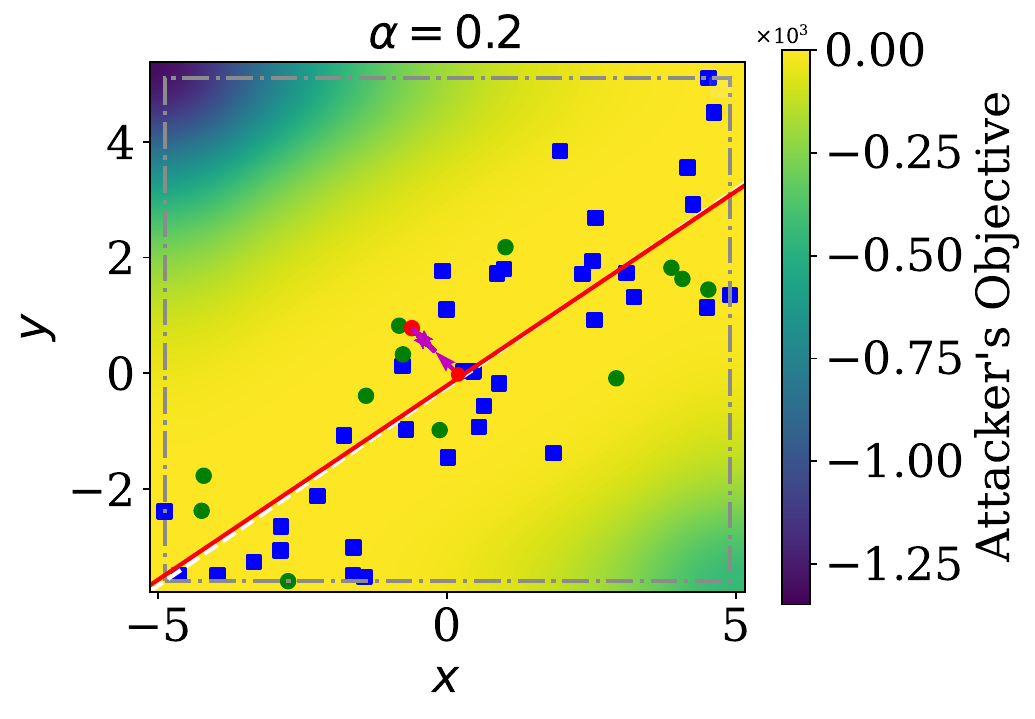}
\label{fig:no_norm_d}\vspace{-0cm}}
\vspace{-.2cm}
\caption{Effect of not normalizing the objective functions in a synthetic example. The blue points are clean points, the green points are poisoning points, and the two red points represent a poisoning point at the beginning and at the end of the optimization process. The magenta line shows the trajectory of this point during optimization. The white dashed line is the regression line learned under the original clean data, and the red solid line is the regression line learned under the complete poisoned training data. The colormap represents the value of the attacker's objective ($\mathcal{A}_d$) as a function of the location of the poisoning point.
(a) $\alpha=1$. (b) $\alpha=0.6$. (c) $\alpha=0.4$. (d) $\alpha=0.2$. Note the sudden transition for the poisoning points from outliers ($\alpha=1$) to inliers ($\alpha=\{0.6,0.4,0.2\}$). This is because the detectability objective dominates the effectiveness objective.
\vspace{-0.3cm}}
\label{fig:no_norm}
\end{figure}

Gradient-based approaches can be very sensitive to different value ranges in the two objectives. In our case, the gradients can have significantly different norms for one objective compared to the other, leading to one objective completely dominating the optimization \cite{milojkovic2019multi}. This phenomenon is shown in Fig.~\ref{fig:no_norm}, where there is a sudden transition from outliers ($\alpha=1$) to inliers ($\alpha=\{0.6, 0.4, 0.2\}$) because the detectability risk is dominant. Our normalization approach mitigates this effect.

Inspired by normalization methods in multiobjective optimization \cite{marler2004survey}, we normalize each objective by their respective maximum value.

The maximum of the detectability-risk function, $\mathcal{R}$, is calculated as the maximum of the attacker's objective in Eq.~(\ref{eqAttacker_}), when the effectiveness objective is zero: $\max_{\mathcal{D}_\mathrm{p}' \in \Phi(\mathcal{D}_\mathrm{p})}  \mathcal{A}_d|_{\mathcal{L}=0}$. Because the parameters are learned on a clean training set, the detectability objective does not depend on previous batches of poisoning points: $\mathcal{R}_\mathrm{norm}\approx  {\mathcal{R}}/{\mathcal{R}_\mathrm{ref}} = \frac{\mathcal{R}}{
\max_{\mathcal{D}_\mathrm{p}' \in \Phi(\mathcal{D}_\mathrm{p})}  \mathcal{A}_d|_{\mathcal{L}=0}}$ and the detectability risk is similar for all batches of poisoning points.

In practice, we calculate the normalization coefficient as the maximum value of the detectability risk across the batches, when $\alpha=1$ as described in Alg.~\ref{alg:norm_obj} (Appx.~\ref{sec:norm_obj}).

\begin{figure}[!t]
\centering
\subfloat[]{\includegraphics[width=1.7in]{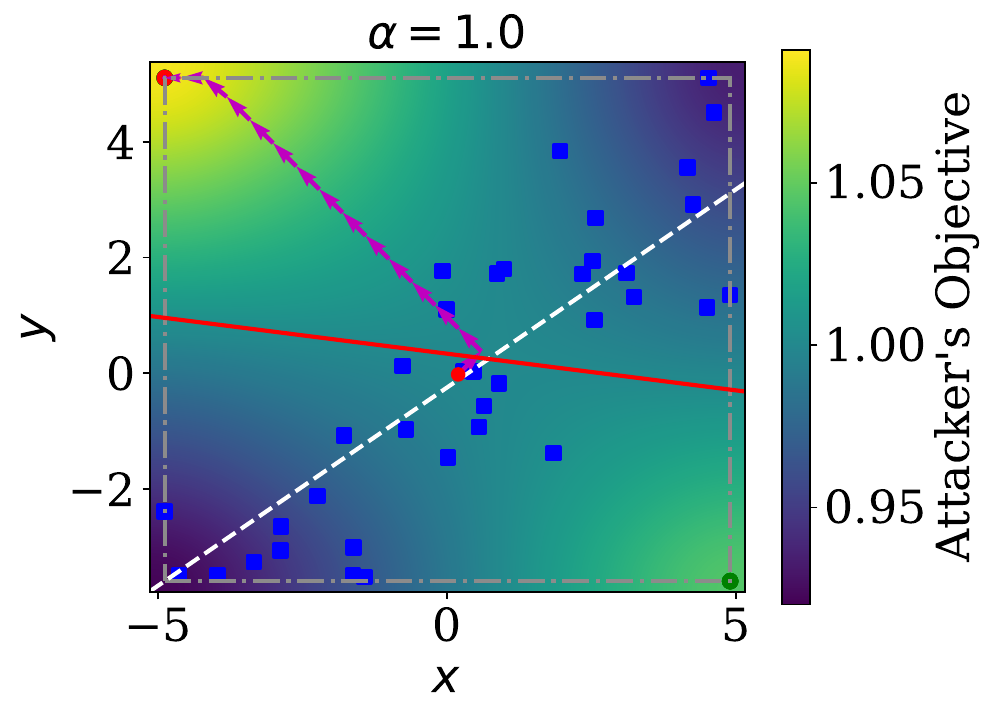}
\label{fig:det_losst_a}}
\subfloat[]{\includegraphics[width=1.6in]{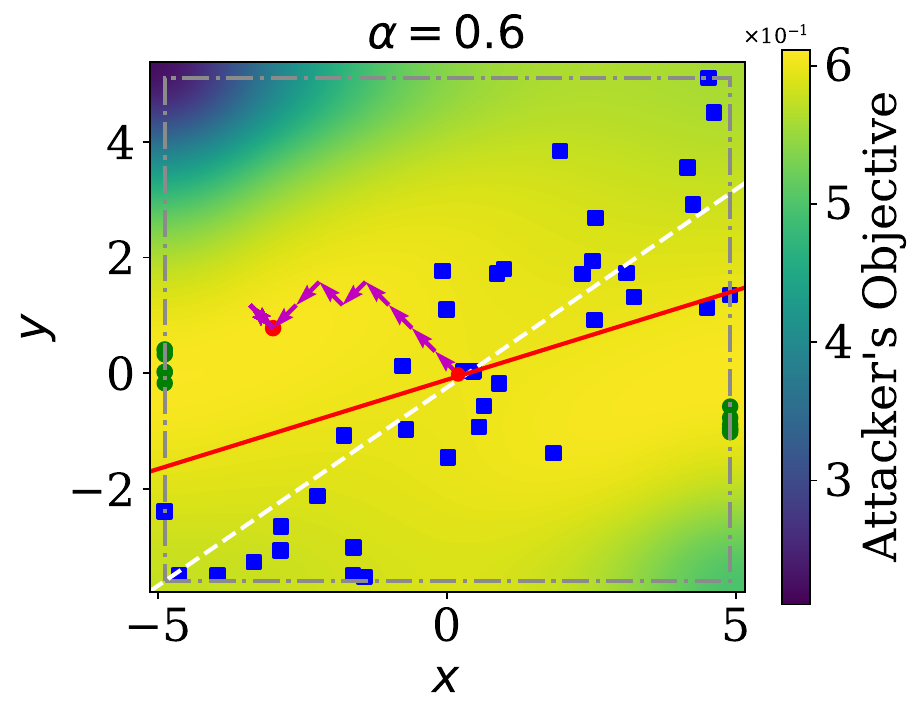}
\label{fig:det_loss_b}}
\\
\vspace{-0.4cm}
\subfloat[]{\includegraphics[width=1.7in]{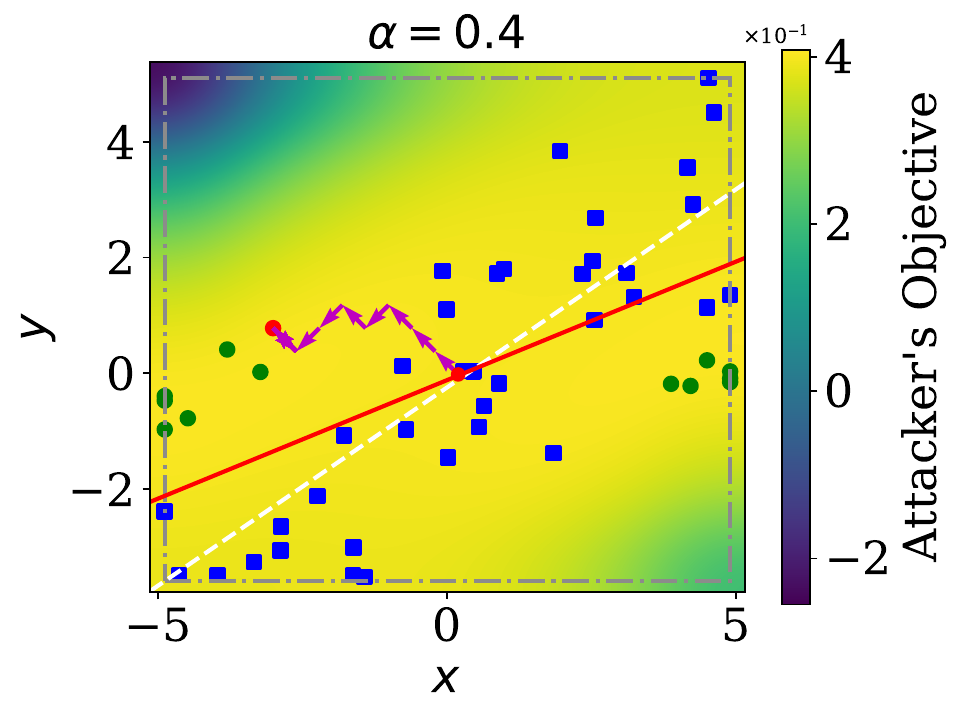}
\label{fig:det_loss_c}}
\subfloat[]{\includegraphics[width=1.7in]{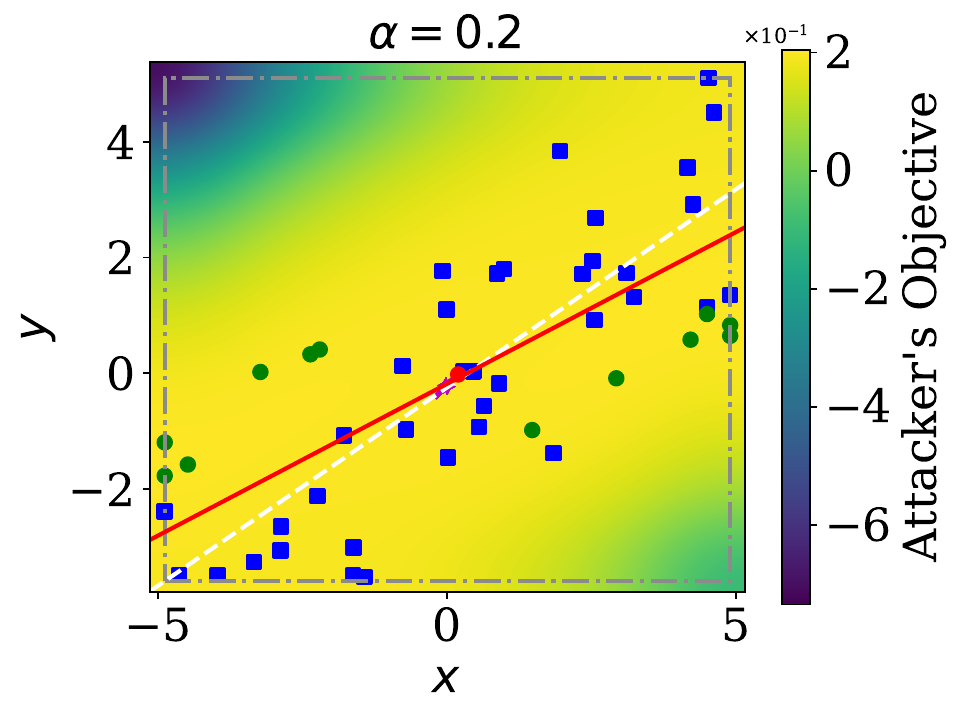}
\label{fig:det_loss_d}}
\vspace{-.2cm}
\caption{Effect of normalizing the objective functions in a synthetic example. The blue points are the clean points, the green points are the poisoning points, and the two red points represent a poisoning point at the beginning and at the end of optimization. The trajectory of this point is shown by the magenta line. The white dashed line is the regression line learned on clean data, and the
red solid line is the regression line learned under poisoning. The colormap represents the value of the attacker's objective ($\mathcal{A}_d$) as a function of the location of the poisoning point.
(a) $\alpha=1$. (b) $\alpha=0.6$. (c) $\alpha=0.4$. (d) $\alpha=0.2$. Compared to Fig.~\ref{fig:no_norm}, normalizing the objective functions allows a smoother transition for the poisoning points from outliers ($\alpha=1$) to inliers ($\alpha=\{0.6,0.4,0.2\}$).\vspace{-0.4cm}}
\label{fig:det_loss__}
\end{figure}

\begin{figure}[!t]
\centering
\subfloat[]{\includegraphics[width=1.5in]{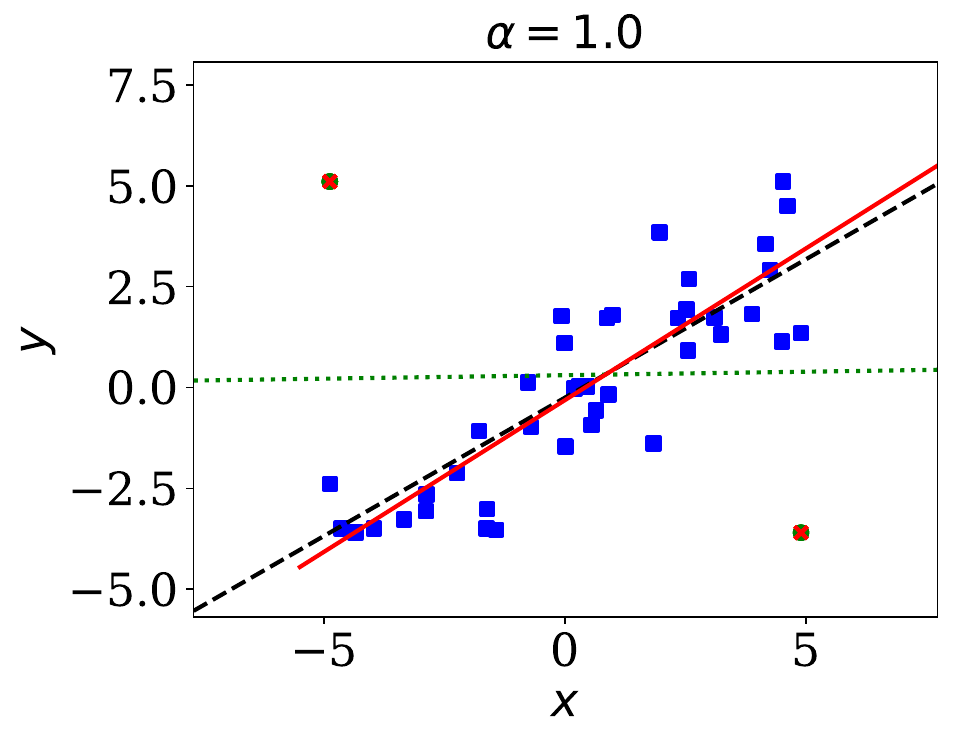}
\label{fig:trim_alpha_a}}
\subfloat[]{\includegraphics[width=1.5in]{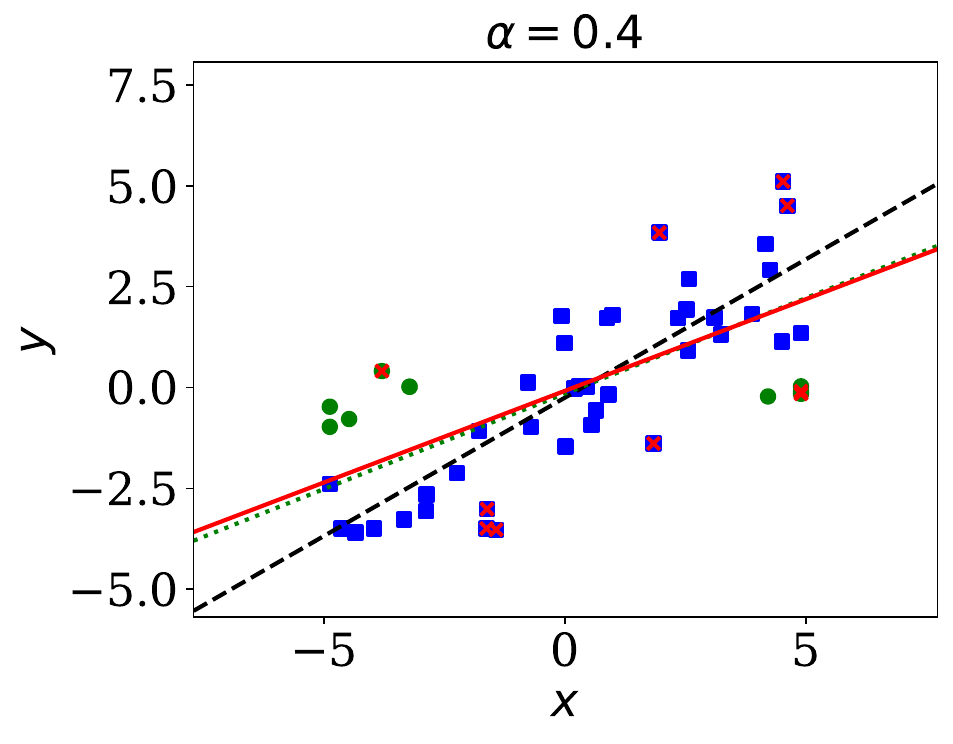}
\label{fig:trim_alpha_b}}
\vspace{-.2cm}
\caption{Synthetic example showing the effect of stealthy poisoning against TRIM \cite{jagielski2018manipulating}, for $\alpha=1$ and $\alpha=0.4$.
The blue points are the clean points, the green points are the poisoning points, the red crosses are points rejected by TRIM, The black dashed line is the regression line learned on clean data, the green dotted line is the regression line learned on the complete poisoned training data, and the red solid line is the regression line learned on training data not rejected by TRIM. TRIM fails to detect stealthy poisoning points, because they are closer to the clean distribution.\vspace{-0.3cm}}
\label{fig:trim_alpha}
\end{figure}

For the effectiveness objective, the normalization coefficient depends on the size of the batch of points, and on whether the objective is measured on the clean or the poisoned model. Therefore, it is difficult to obtain the maximum value of the objective, $
\max_{\mathcal{D}_\mathrm{p}' \in \Phi(\mathcal{D}_\mathrm{p})}  \mathcal{L}(\mathcal{D}_\mathrm{val},    {\bf w}^\star)|_{\mathcal{R}=0}$, as this value depends on the values of the previous batches of poisoning points, which may, in turn, be constrained. To address this challenge, we calculate a new normalization coefficient for each batch i.e., for the $i$th batch, we re-normalize the effectiveness objective as follows: $\mathcal{L}_\mathrm{norm}\left(\mathcal{D}_\mathrm{val},    {\bf w}^{\star^{(i)}}\right)$ $\approx { \mathcal{L}\left(\mathcal{D}_\mathrm{val},    {\bf w}^{\star^{(i)}}\right)}/{\mathcal{L}_\mathrm{ref}}$ $\approx \frac{ \mathcal{L}\left(\mathcal{D}_\mathrm{val},    {\bf w}^{\star^{(i)}}\right)}{ \mathcal{L}\left(\mathcal{D}_\mathrm{val},    {\bf w}^{\star^{(i-1)}}\right)}$, where ${\bf w}^{\star^{(i-1)}}$ are the poisoned parameters trained on the previous optimal poisoned set. When computing the first batch, the previous poisoned set is the clean training set, and therefore ${\bf w}^{\star^{(i-1)}}={\bf w}^{\star}_{\mathrm{tr}}$. Although this heuristic does not calculate the maximum of the objective, it provides a good empirical reference for the range of values the effectiveness objective can take. Different values of $\alpha$ lead to different normalization coefficients see Alg.~\ref{alg:pois2} (Appx.~\ref{sec:norm_obj}).
e
In summary, to solve the multiobjective bilevel optimization problem (\ref{eqAttacker_}), we first compute the normalization coefficient for the detectability-risk function for $\alpha = 1$ (Alg.~\ref{alg:norm_obj}), and then, for  $\alpha \neq 1$, we compute the corresponding normalization coefficient for the effectiveness function (Alg.~\ref{alg:pois2}). Fig.~\ref{fig:det_loss__} shows the effect of normalizing the objective functions in a synthetic example. Note that normalization enables a smoother transition as the value of $\alpha$ changes. Fig.~\ref{fig:trim_alpha} shows that poisoning points close to the clean distribution ($\alpha=0.4$) shift the regression line even when TRIM is deployed as a defense. In contrast, poisoning points without detectability constraints ($\alpha=1$) are easily filtered by TRIM.

\subsection{Solving Optimal Poisoning Attacks with Soft Detectability Constraints}
\label{subsec:genpois2}

Solving Eq.~(\ref{eqAttacker_}) is strongly NP-Hard \cite{bard2013practical}. Even when the inner problem is convex, the bilevel problem is, in general, non-convex. However, gradient-based approaches can be used to find local optima i.e., (possibly) suboptimal solutions.

State-of-the art attacks against regression algorithms \cite{jagielski2018manipulating, wen2021great} are based on the implicit function theorem, and thus require training the whole learning algorithm to compute the hypergradient (gradient of the attacker's objective function), i.e., until the stationarity conditions are met. To sidestep this problem, different techniques are used to estimate the hypergradients \cite{domke2012generic,maclaurin2015gradient,pedregosa2016hyperparameter,franceschi2017forward,munoz2017towards,franceschi2018bilevel, grazzi2020iteration} without completely retraining each time the hypergradient is computed. Instead, they estimate the hypergradient by truncating learning in the inner problem to a reduced number of iterations.

We include in Alg.~\ref{alg:bg2} (Appx.~\ref{sec:rmd_mult})
the Reverse-Mode Differentiation (RMD) algorithm  we use to compute the hypergradient estimate in the outer level problem (both for the features of the poisoning points, ${\bf X}_\mathrm{p}$, and their labels, ${\bf y}_\mathrm{p}$). For more details, including the algorithm complexity of RMD, see \cite{franceschi2017forward, franceschi2018bilevel, grazzi2020iteration, carnerero2024hyperparameter}.

After computing the hypergradients, at each \emph{hyperiteration} we use projected hypergradient ascent to update the poisoning points and the hyperparameters:

\begin{equation}
  \begin{aligned}
{\bf X}_\mathrm{p}  \leftarrow \Pi_{\Phi(\mathcal{D}_\mathrm{p})} ( {\bf X}_\mathrm{p} + \gamma \ \nabla_{{\bf X}_\mathrm{p}} \mathcal{A}_d ), \\
{\bf y}_\mathrm{p}  \leftarrow \Pi_{\Phi(\mathcal{D}_\mathrm{p})} ( {\bf y}_\mathrm{p} + \gamma \ \nabla_{{\bf y}_\mathrm{p}} \mathcal{A}_d ),
  \end{aligned}
\label{eqUpdates_}
\end{equation} where $\gamma$ is the learning rate for the outer problem and $\Pi_{\Phi(\mathcal{D}_\mathrm{p})}$ is the projection operator for the poisoning points, $\mathcal{D}_\mathrm{p}$, so that their updated values are within their feasible domain, $\Phi(\mathcal{D}_\mathrm{p})$. These projection operators are defined as $\Pi_{\Phi{(\cdot)}} (input) 	\triangleq \mathrm{clip}(input, \inf \Phi(\cdot)$, $ \sup \Phi(\cdot))$, so that their values remain within the corresponding feasible domains, $\Phi(\cdot)$. The training procedure is summarized in Alg.~\ref{alg:hyperreg_} (Appx.~\ref{sec:proj}).

\subsection{Experimental Evaluation of Stealthy Attacks}
\label{sec:expreg}

We now empirically show that our attack formulation with soft detectability constraints bypasses state-of-the-art defenses. For this, we  evaluate the effectiveness and stealth of poisoning  attacks in stationary settings. Thus, we perform a random split of the datasets to simulate absence of concept drift.\footnote{A time-aware split, rather than a random split would make it difficult to determine whether the effectiveness of the poisoning attack is due to an weakness of the regression model or due to the natural evolution of data \cite{pendlebury2019tesseract}.} We use four real-world datasets: interest rate of loans (\emph{Loan})---made on the Lending Club peer-to-peer lending platform \cite{kan2013lending} (preprocessed as in \cite{jagielski2018manipulating});
heart disease (\emph{Heart Disease}) \cite{weerasinghe2021closing}; Boston housing prediction (\emph{Boston Housing}) \cite{harrison1978hedonic}; and appliances energy prediction  (\emph{Appliances}) \cite{candanedo2017lappliances} (preprocessed as in \cite{weerasinghe2021closing}). These are also the datasets used in the related work on poisoning attacks or in standard regression benchmarks. All datasets are drawn at random from the original joint pool of points and details for each dataset are included in Table~\ref{tabDatasets2} (Appx.~\ref{subsec:expset2}).

We consider indiscriminate attacks on LR and DNNs. For LR, this is in line with the related work on poisoning regression models \cite{mei2015using, zhang2018training, jagielski2018manipulating, muller2020data, wen2021great, weerasinghe2021closing, li2021optimal}.
We evaluate defenses in the context for which they have been proposed. Although most previous work on optimal poisoning attacks against regression models focuses solely on linear models, we also evaluate our attack strategy in Eq.~(\ref{eqAttacker_}) against feed-forward DNNs. We provide an analysis of the robustness of the learning algorithms to worst-case scenarios for attacks with different strength and detectability constraints.

When presenting our results, we show both the test Normalized Mean
Squared Error (NMSE) of the model when no defense has been deployed and the \emph{test defense gain} of a particular defense as a function of the number of poisoning points. The NMSE is computed as:
\begin{equation}
    \text{NMSE}_\text{test, no defense} = {\mathrm{MSE}_\text{test, no defense}}/\left({||{\bf y}_\text{test}||_2^2/n_\text{test}}\right),
\end{equation}
where $n_\text{test}$ is the number of test samples and ${\bf y}_\text{test}$ are the test labels. The \emph{test defense gain} of defense $i$, is defined as:
\begin{equation}
    \text{Test Defense Gain}_i [\%] = \frac{\mathrm{NMSE}_\text{test, no defense} - \mathrm{NMSE}_\mathrm{{test}, i}}{\mathrm{NMSE}_\text{test, no defense}}\cdot 100,
\end{equation}
This test defense gain quantifies the increase or decrease in test MSE when a defense is deployed. A negative gain means that the defense worsens model performance.

We evaluate our attack against several benchmark defenses: TRIM \cite{jagielski2018manipulating}, Huber \cite{huber1964robust}, and SEVER \cite{diakonikolas2019sever} testing attacks with different levels of detectability, i.e., for different values of $\alpha$. We examine the trade-off between effectiveness and stealthiness: $\alpha=1$ are attacks entirely dominated by effectiveness, $\alpha=0.3$ are stealthy attacks that achieve some effectiveness, and $\alpha=0.1$ are attacks dominated by stealthiness.

In each plot, the total defense gain with respect to the base-case (no-defense) scenario is reported as a function of the fraction of poisoning points. A negative test defense gain (the area shadowed in red) shows that the defense is less effective than the unprotected model, i.e., \textit{the defense is not worth deploying}. A positive gain shows that the defense mitigates the attack to some extent. More detailed experimental settings can be found in Appx.~\ref{subsec:expset2}.

\subsubsection{Linear Regression}

\begin{figure}[!t]
\centering
\subfloat[]{\includegraphics[width=1.5in]{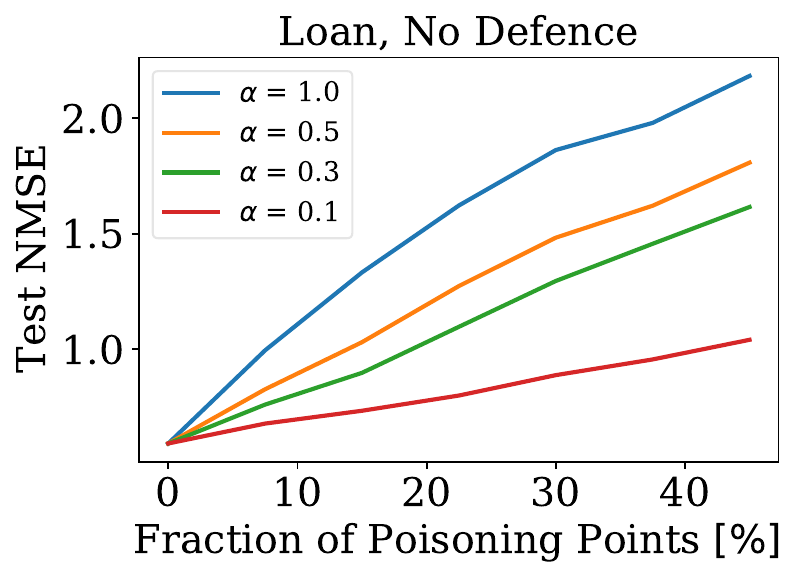}
\label{fig:mse_tst_alphas_a}}
\subfloat[]{\includegraphics[width=1.4in]{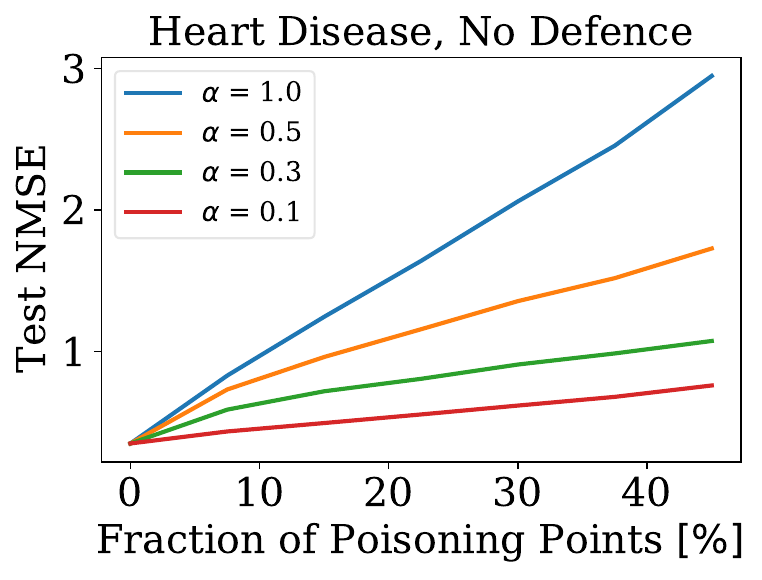}
\label{fig:mse_tst_alphas_b}}
\vspace{-.2cm}
\caption{Test NMSE of LR when there is no defense deployed, for $\alpha=1$,  $\alpha=0.5$,  $\alpha=0.3$, and  $\alpha=0.1$.  (a) Loan. (b) Heart Disease.
\vspace{-0.5cm}}
\label{fig:mse_tst_alphas}
\end{figure}

We first report in Fig.~\ref{fig:mse_tst_alphas} (and also in Appx.~\ref{subsec:addit_results_lr}, Fig.~\ref{fig:mse_tst_alphas2})
the results
corresponding to the \emph{base case} without defenses. We observe that the test NMSE increases with the value of $\alpha$, as we generate more points that are out of distribution. In contrast, reducing $\alpha$ constrains the poisoning points, with the goal of bypassing defenses, as we see next.

\begin{figure}[!t]
\centering
\subfloat[]{\includegraphics[width=1.5in]{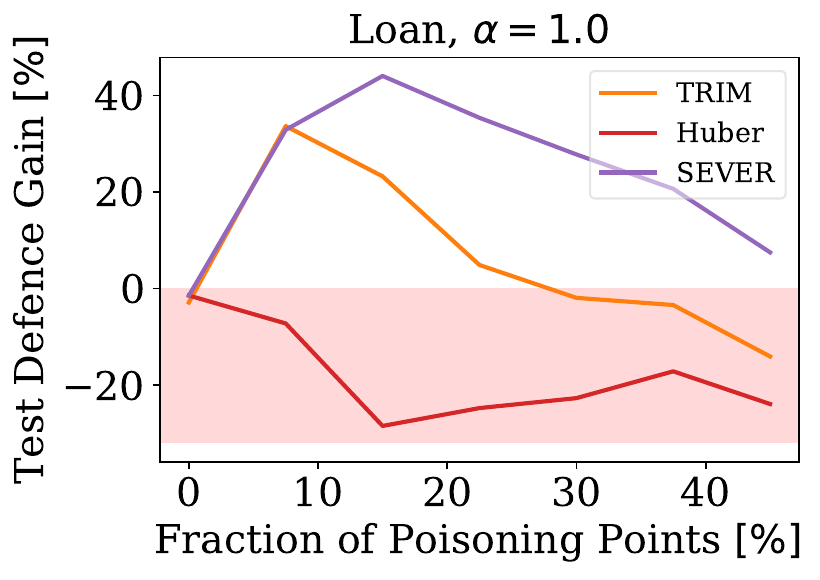}
\label{fig:mse_tst_a}}
\subfloat[]{\includegraphics[width=1.5in]{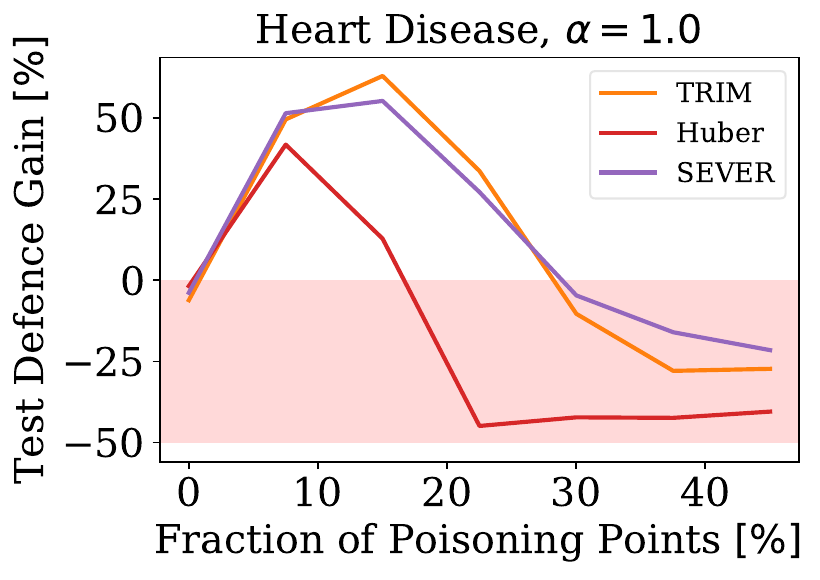}
\label{fig:mse_tst_b}}
\\
\vspace{-0.4cm}
\subfloat[]{\includegraphics[width=1.5in]{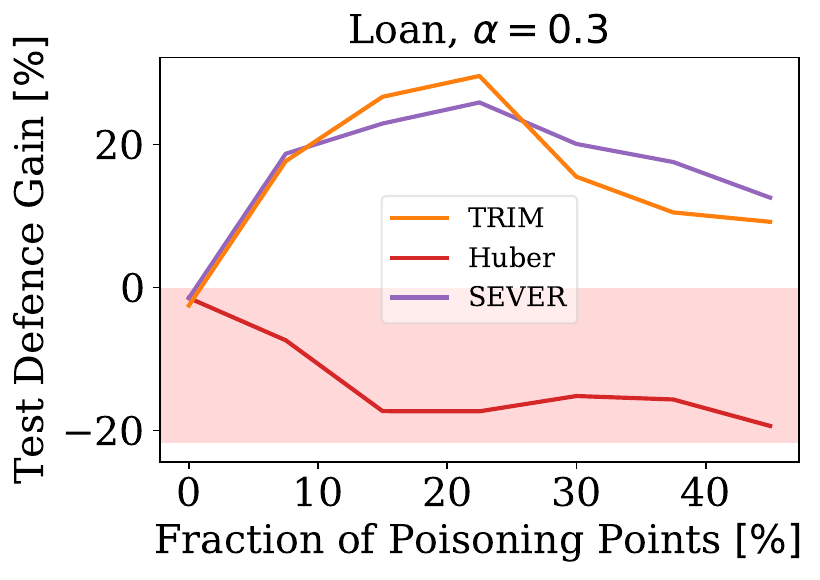}
\label{fig:mse_tst_c}}
\subfloat[]{\includegraphics[width=1.5in]{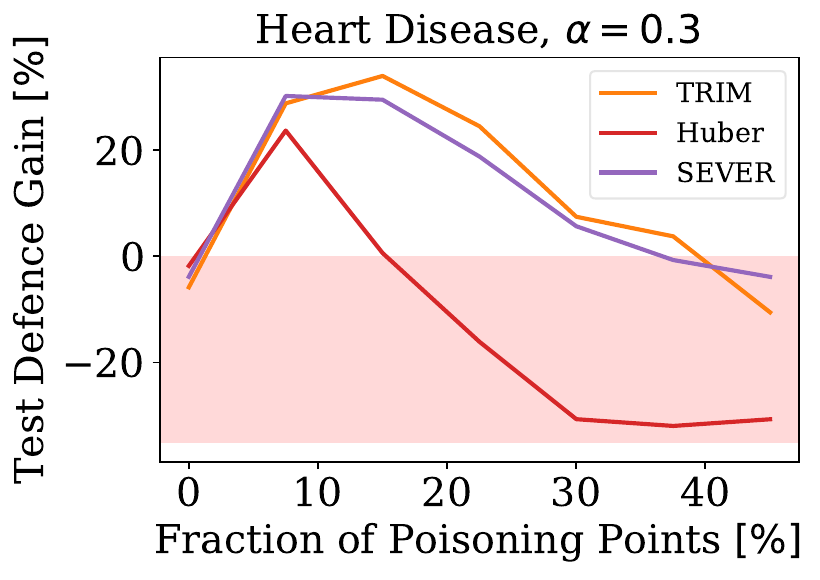}
\label{fig:mse_tst_d}}
\\
\vspace{-0.4cm}
\subfloat[]{\includegraphics[width=1.5in]{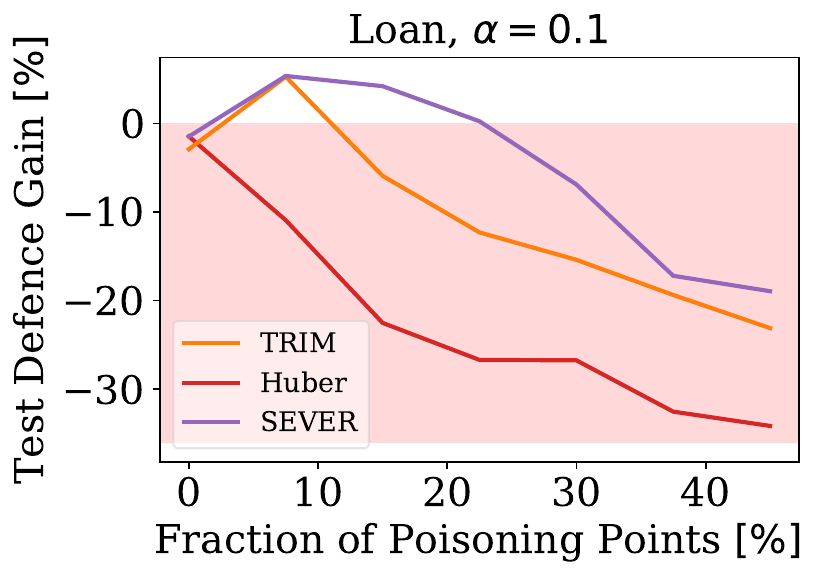}
\label{fig:mse_tst_e}}
\subfloat[]{\includegraphics[width=1.5in]{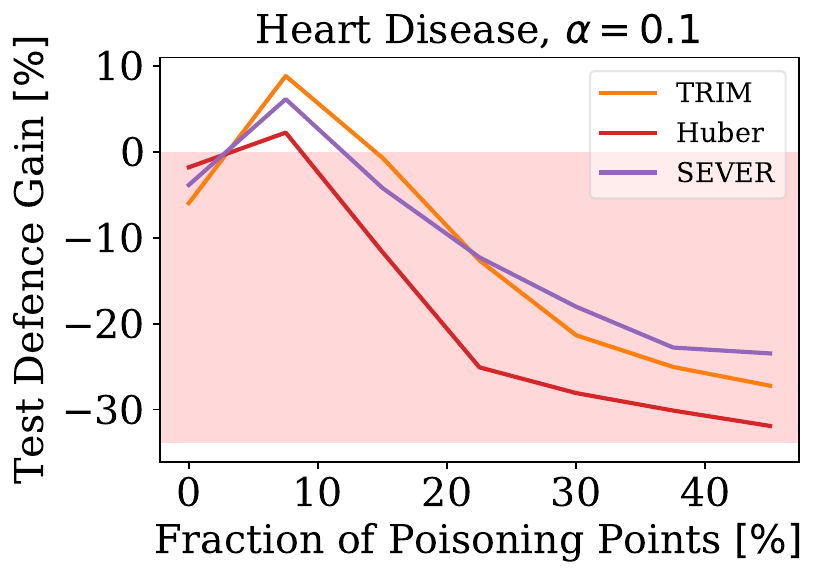}
\label{fig:mse_tst_f}}
\vspace{-.2cm}
\caption{Test defense gain of LR when using TRIM, Huber, and SEVER. The first and second columns correspond to Loan and Heart Disease, and the first, second, and third rows correspond to $\alpha = 1$, $\alpha = 0.3$, and $\alpha = 0.1$, correspondingly.\vspace{-0.4cm}}
\label{fig:mse_tst}
\end{figure}

\begin{figure}[!t]
\centering
\subfloat[]{\includegraphics[width=1.6in]{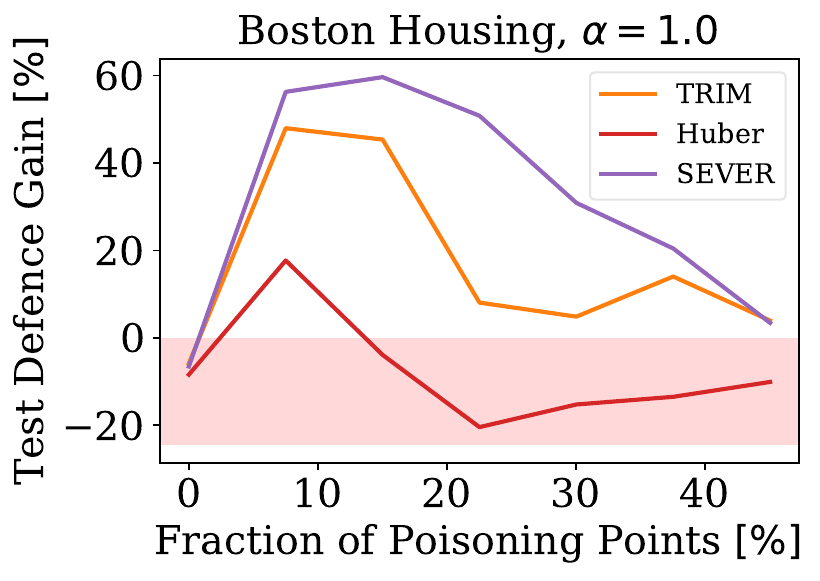}
\label{fig:mse_tst_g}}
\subfloat[]{\includegraphics[width=1.6in]{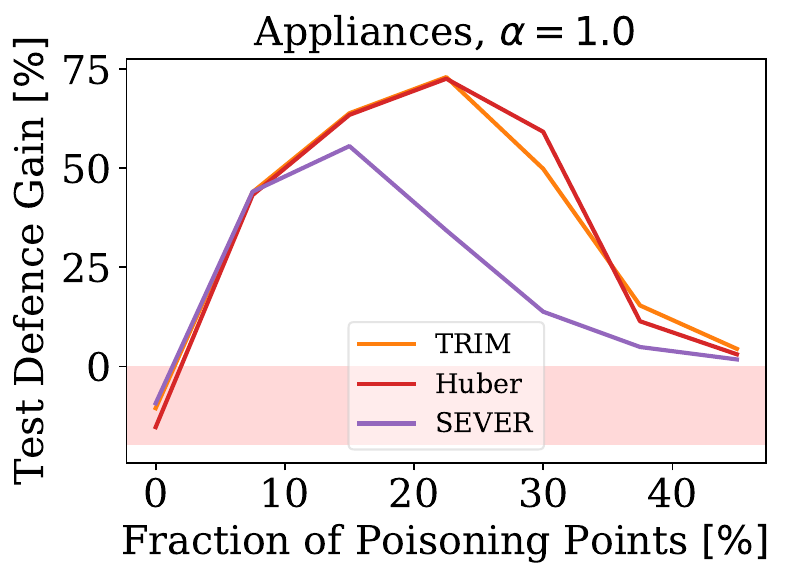}
\label{fig:mse_tst_h}}
\\
\vspace{-0.4cm}
\subfloat[]{\includegraphics[width=1.6in]{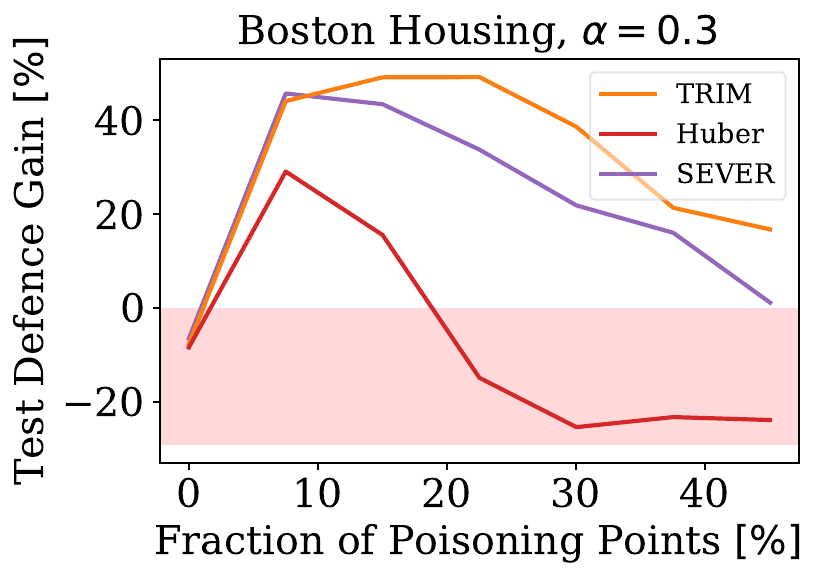}
\label{fig:mse_tst_i}}
\subfloat[]{\includegraphics[width=1.6in]{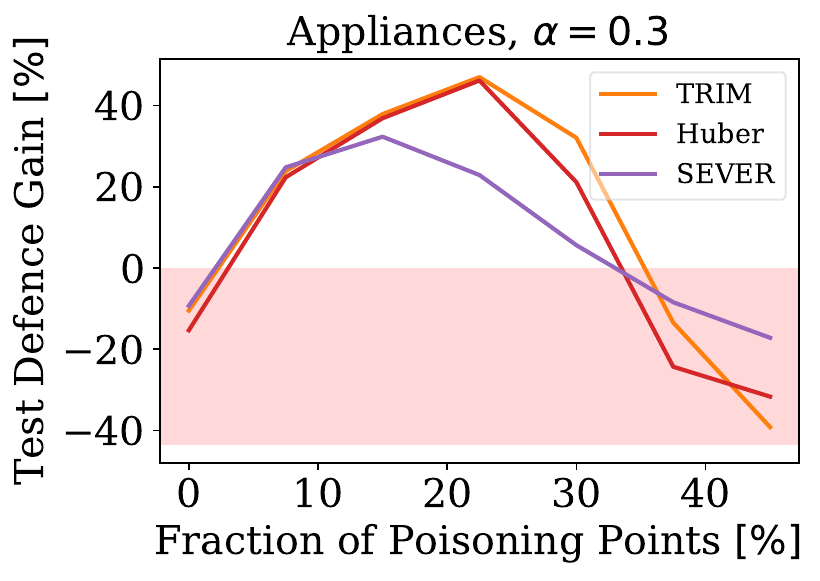}
\label{fig:mse_tst_j}}
\\
\vspace{-0.4cm}
\subfloat[]{\includegraphics[width=1.6in]{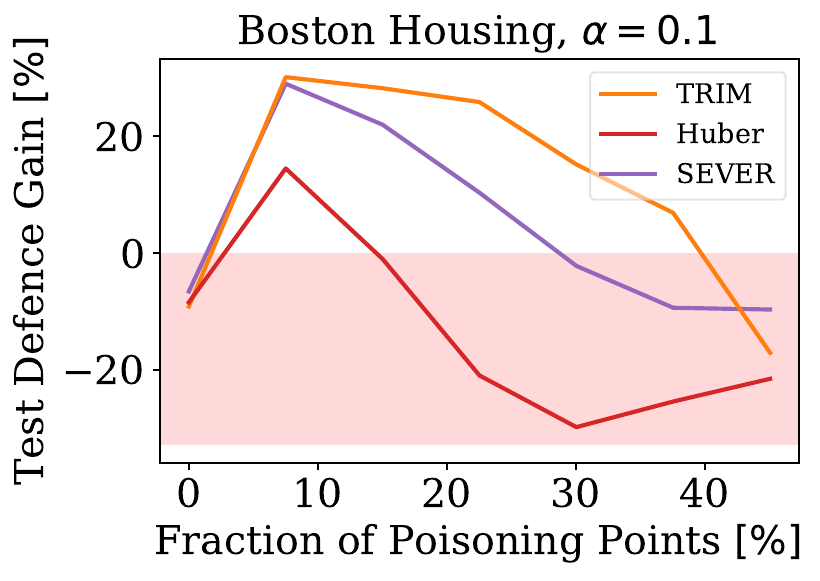}
\label{fig:mse_tst_k}}
\subfloat[]{\includegraphics[width=1.6in]{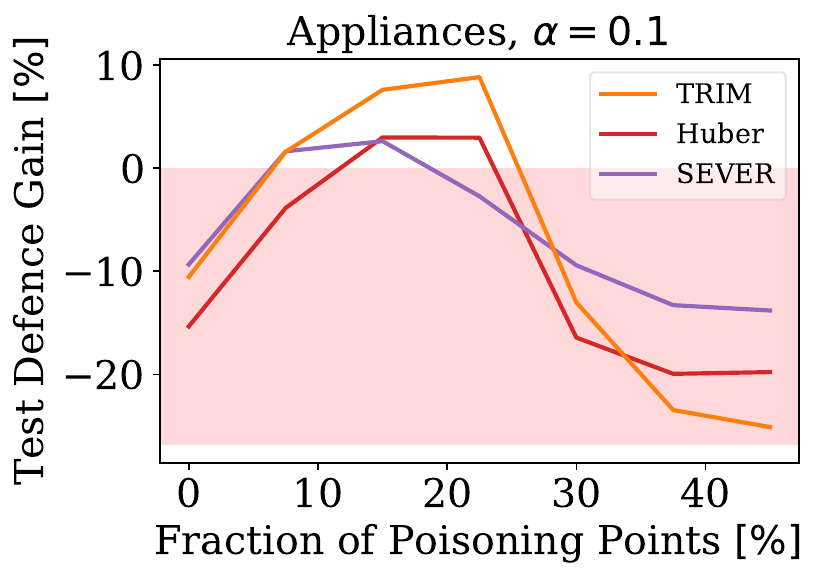}
\label{fig:mse_tst_l}}
\vspace{-.2cm}
\caption{Test defense gain of LR when using TRIM, Huber, and SEVER. The first and second columns correspond to Boston Housing and Appliances, respectively. The first, second, and third rows correspond to $\alpha = 1$, $\alpha = 0.3$, and $\alpha = 0.1$, correspondingly.\vspace{-.3cm}}
\label{fig:mse_tst2}
\end{figure}

Fig.~\ref{fig:mse_tst}~and~\ref{fig:mse_tst2} illustrate the performance of the attack against benchmark defenses. Overall, these results show that current defenses protect poorly against stealthy poisoning attacks.
In particular, Huber performs poorly for different values of $\alpha$ for all data sets, except Appliances. For $\alpha=1$ attacks are entirely dominated by effectiveness, yet some defenses fail and actually increase the test NMSE w.r.t. the no-defense scenario. They are beneficial only for low fractions of poisoning points (e.g., below $15\%$) and have a damaging effect in all other cases.
For most defenses, when the attack is entirely stealthy ($\alpha = 0.1$), the test defense gain is negative i.e., \textit{using the defense is more damaging than not using any defense at all}.

\subsubsection{Deep Neural Networks}

We also target feed-forward DNNs. To our knowledge, this is the first study that analyzes optimal data poisoning attacks against DNNs in regression settings. This also demonstrates the scalability of our approach as the computation is more challenging in this setting.

\begin{figure}[!t]
\centering
\subfloat[]{\includegraphics[width=1.55in]{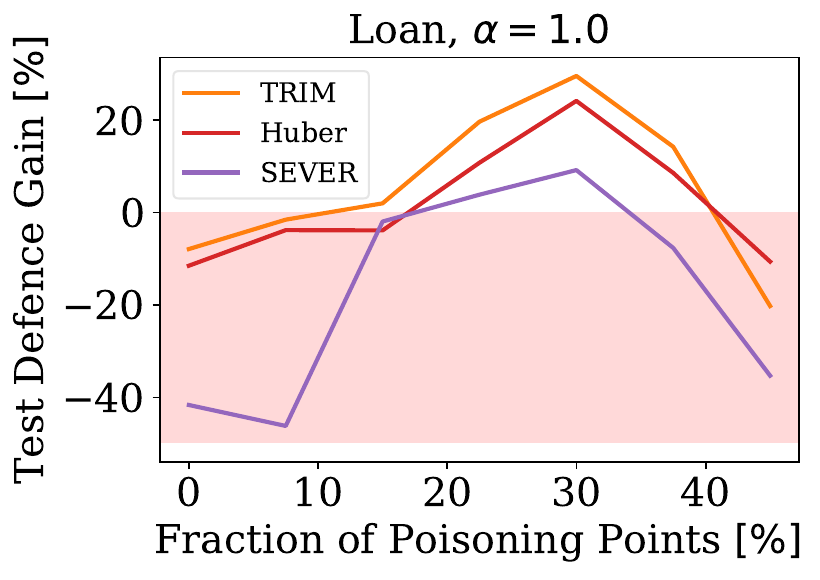}
\label{fig:mse_tst_dnn_no_a}}
\subfloat[]{\includegraphics[width=1.55in]{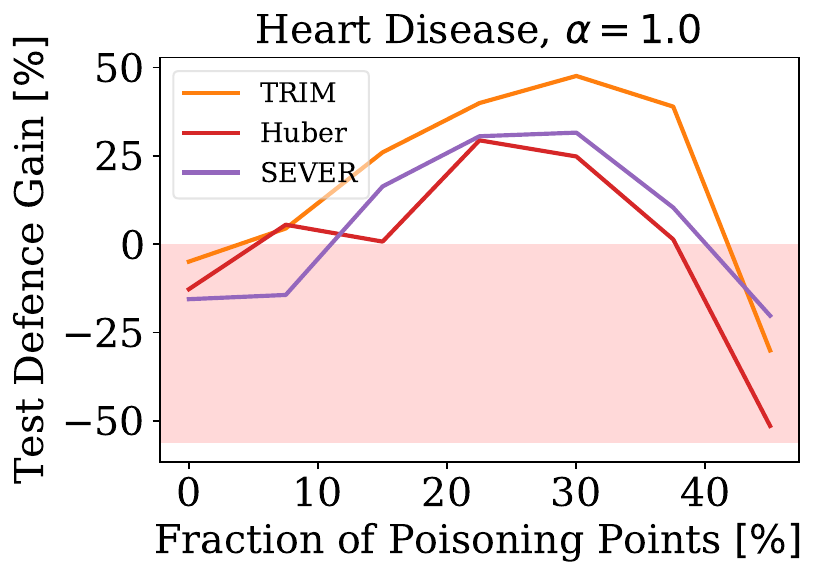}
\label{fig:mse_tst_dnn_no_d}}
\\
\vspace{-0.4cm}
\subfloat[]{\includegraphics[width=1.55in]{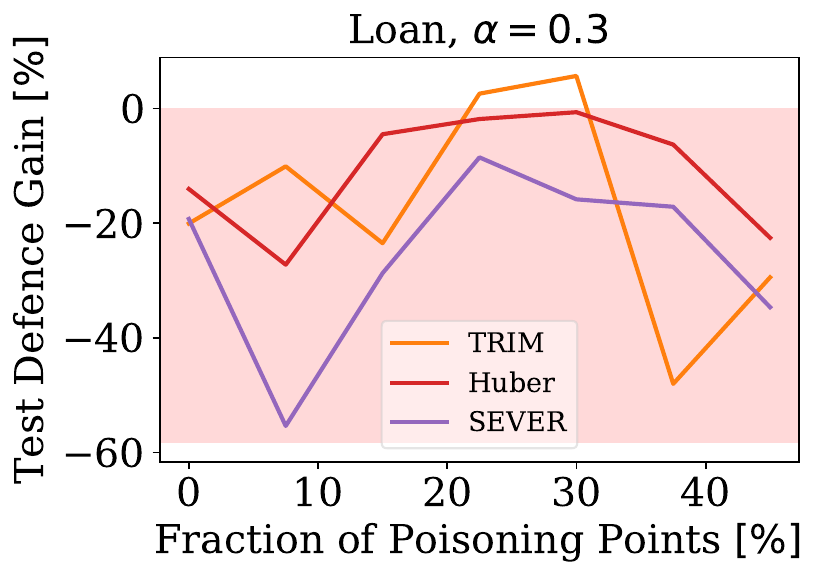}
\label{fig:mse_tst_dnn_no_b}}
\subfloat[]{\includegraphics[width=1.55in]{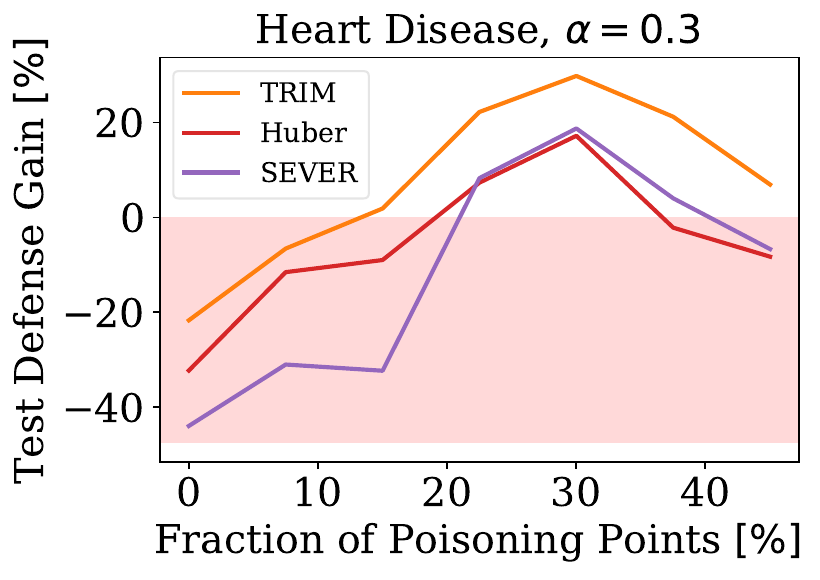}
\label{fig:mse_tst_dnn_no_e}}
\\
\vspace{-0.4cm}
\subfloat[]{\includegraphics[width=1.55in]{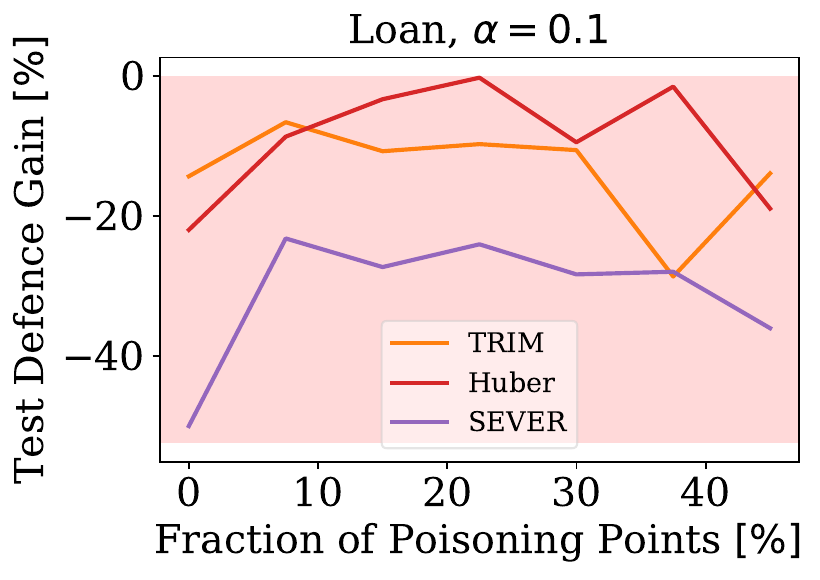}
\label{fig:mse_tst_dnn_no_c}}
\subfloat[]{\includegraphics[width=1.55in]{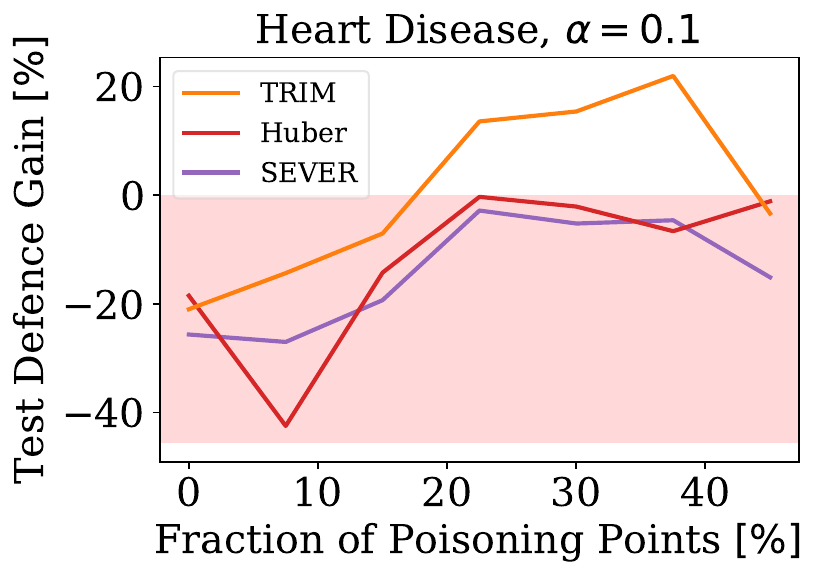}
\label{fig:mse_tst_dnn_no_f}}
\vspace{-.2cm}
\caption{Test defense gain of the DNNs when using TRIM, Huber, and SEVER. The first and second columns correspond to Loan and Heart Disease, respectively. The first, second, and third row correspond to $\alpha = 1$, $\alpha = 0.3$, and $\alpha = 0.1$, correspondingly.\vspace{-.5cm}}
\label{fig:mse_tst_dnn_no}
\end{figure}

Fig.~\ref{fig:mse_tst_dnn_no} illustrates the performance of the attack against DNNs when considering TRIM, Huber, and SEVER, similar to the experiments shown for LR. The takeaway is that in the case of DNNs as well, most defenses fall short when the attacker considers detectability. For the Loan dataset, comparing Fig.~\ref{fig:mse_tst_dnn_no} and Fig.~\ref{fig:mse_tst} suggests that, overall, the defenses perform worse in the case of the DNNs, as the test defense gain is lower.

\section{\emph{BayesClean}: A Defense Based on Uncertainty}
\label{sec:bayesclean}

Our results above have shown that state-of-the-art defenses fail when poisoning points are constrained reducing the distinction between clean and poisoning distributions.
As the number of poisoning points increases, there is a breaking point where the reference for the true distribution is too poisoned to be useful. Then, we can no longer reject points solely based on their errors as the errors of the clean points can be larger than those of the poisoning points. Moreover, some defenses \cite{jagielski2018manipulating,zhang2018training, diakonikolas2019sever, wen2021great} assume an ``a-priori'' ratio of poisoning points to reject. This ratio is difficult to estimate in practice and leads to even more genuine points being rejected.

To overcome these limitations, we develop a novel defense based on Bayesian LR, which we call \emph{BayesClean}. This defense does not assume a ratio of poisoning points to reject, and is more robust when poisoning points are constrained but more numerous. The intuition is to reject poisoning points outside of a ``confidence interval'' around the mean prediction of the Bayesian LR model. We first describe the fundamentals of this model.

\subsection{Bayesian Linear Regression}

As with standard LR, Bayesian LR models the target as
$y = {\bf x}^{\textsf T}{\bf w} + \varepsilon$,
with Gaussian noise $\varepsilon\!\sim\!\mathcal N(0,\beta^{-1})$, where
$\beta$ is the noise precision. A Gaussian prior is placed on the weights,
$p({\bf w}|\lambda)=\mathcal N({\bf 0},(\lambda^{-1}{\bf I}_d))$,
with precision $\lambda$.
Following \cite{tipping2001sparse}, both $\lambda$ and $\beta$ have Gamma
hyperpriors with small parameters
($\lambda_1=\lambda_2=\beta_1=\beta_2=10^{-6}$)
to ensure non-informativeness.

Given data $({\bf X},{\bf y})$, the posterior factorizes as
$p({\bf w},\lambda,\beta|{\bf X},{\bf y})
=p({\bf w}|{\bf X},{\bf y},\lambda,\beta)\,p(\lambda,\beta|{\bf X},{\bf y})$.
The weight posterior is Gaussian \cite{bishop2006pattern}: $p({\bf w}|{\bf X},{\bf y},\lambda,\beta)
=\mathcal N({\boldsymbol\mu}_{\text{post}},
{\boldsymbol\Sigma}_{\text{post}})$, ${\boldsymbol\Sigma}_{\text{post}}
=(\beta{\bf X}^{\textsf T}{\bf X}+\lambda{\bf I}_d)^{-1}, $${\boldsymbol\mu}_{\text{post}}
=\beta{\boldsymbol\Sigma}_{\text{post}}{\bf X}^{\textsf T}{\bf y}$. Under the \textit{evidence approximation}
\cite{gull1989developments,mackay1992bayesian,tipping2001sparse},
$p(\lambda,\beta|{\bf X},{\bf y})$ is assumed sharply peaked at
$(\hat\lambda,\hat\beta)$. Then, learning
the hyperparameters becomes the search for the mode of the
hyperparameters’ log-posterior:

$\max_{\lambda, \beta}\log p(\lambda,\beta|{\bf X},{\bf y})
\propto \log p({\bf y}|{\bf X},\lambda,\beta)
+\log p(\lambda)+\log p(\beta)$,

typically solved via an Expectation-Maximization (EM) procedure \cite{mackay1992bayesian, tipping2001sparse}. Predictions for a new input ${\bf x}_*$ follow the predictive distribution
$p(y_*|{\bf x}_*,{\bf X},{\bf y},\hat\lambda,\hat\beta)
=\mathcal N(y_*|\mu_*,\sigma_*^2)$, where

$\mu_*={\boldsymbol\mu}_{\text{post}}^{\textsf T}{\bf x}_*$ and $
\sigma_*^2=\hat\beta^{-1}+{\bf x}_*^{\textsf T}
{\boldsymbol\Sigma}_{\text{post}}{\bf x}_*.$

Here $\hat\beta^{-1}$ represents data noise and
${\bf x}_*^{\textsf T}{\boldsymbol\Sigma}_{\text{post}}{\bf x}_*$
captures model uncertainty.

\subsection{Predictive Variance under Data Poisoning}
\label{subsec:predvar}

In the case when $\hat{\beta}$ is not learned from the data but is set as a constant, as additional data points are observed, the posterior distribution becomes narrower. As a consequence, it can be shown \cite{qazaz1997upper} that the predictive variance becomes smaller, i.e., $\sigma^2_{*n +1} \leq \sigma_{*n}^2$. In the limit where the number of training points is very large, $n \rightarrow \infty$, ${\bf x}_*^{\textsf{T}} {\boldsymbol \Sigma}_\text{post} {\bf x}_*$ goes to zero, and the variance of the predictive distribution arises solely from the additive noise governed by the hyperparameter $\hat{\beta}$. However, in data poisoning settings, the poisoning points do not follow the genuine data distribution. Thus, when $\beta$ is learned from the data, this result does not hold. To see this, consider the expression for the $i$th update of $\beta^{-1}$ in the EM procedure \cite{tipping2001sparse}: $\beta^{{-1}^{(i)}}=\left(\left|\left|{\bf y} - {\bf X}{\boldsymbol \mu}_\text{post}^{(i)}\right|\right|_2^2 + 2 \beta_2\right) \Bigg/ \left(n - \sum_{j=1}^d \gamma_j^{(i)} + 2\beta_1\right)$, where  $\gamma_j^{(i)}\in[0,1]$---also updated in the EM algorithm---can be interpreted as a measure of how ``well-determined'' its corresponding parameter $w_j$ is by the data \cite{mackay1992bayesian}. Thus, if the poisoning points follow a distribution different from the clean distribution, the error term $\left|\left|{\bf y} - {\bf X}{\boldsymbol \mu}_\text{post}^{(i)}\right|\right|_2^2$ can increase, which implies a larger $\beta^{-1}$ and thus a larger predictive variance. This observation leads us to our defense mechanism.

\subsection{\emph{BayesClean}}

BayesClean builds upon a Bayesian LR model, as described in Alg.~\ref{alg:bayesclean} (Appx.~\ref{sec:bayesclean_alg}). First, we learn the optimal hyperparameters $\hat{\lambda}$ and $\hat{\beta}$, where $T_\text{EM}$ is the maximum number of iterations of the EM algorithm. Then, we update the posterior distribution, and compute the posterior predictive distribution for each input point, $\mathcal{D}= ({\bf X}, {\bf y}) =\{({\bf x}_i, y_i)\}_{i=1}^n$. Subsequently, we split $\mathcal{D}$ into three disjoint sets, according to their location around the predictive mean. Points closest to the predictive mean, whose indices we define as $\mathcal{I}_1 = \{i \ | \ |y_i| \leq |\mu_{*_i} + c_1\sigma_{*_i}|\} $, are detected as benign. Points corresponding to the set $\mathcal{I}_2 = \{i \ | \  |\mu_{*_i} + c_1\sigma_{*_i}| < |y_i| \leq |\mu_{*_i} + c_2\sigma_{*_i}|\}$, are flagged for inspection, because they can damage the performance of the model. In our experimental evaluation, we reject these points. Finally, points that lie far from the predictive mean, i.e., $\mathcal{I}_3 = \{i \ | \ |y_i| > |\mu_{*_i} + c_2\sigma_{*_i}|\} $, are rejected. Fig.~\ref{fig:bayesclean} in Appx.~\ref{subsec:predvar_} shows the intuition of BayesClean, and how the predictive variance increases with the ratio of poisoning, as explained in $\S$~\ref{subsec:predvar}. Intuitively, the mismatch between the clean and the poisoning distributions increases the variance and thus the uncertainty in the predictions.

Two advantages of our defense are that \textit{we do not have to define in advance the number of points to reject} (this is impossible to estimate in practice-as the attack is unknown) and that \textit{we do not require a trusted set}.
While constants $c_2\geq c_1$ are defined in advance, we keep their values fixed for any poisoning ratio and across different datasets.

Concerning the complexity of the algorithm, the updates of the hyperparameters depend on computing the covariance matrix of the parameters' posterior, which requires an inverse operation. In practice, we leverage the Singular Value Decomposition (SVD) of the design matrix, ${\bf X}$.
Efficient numerical methods scale in time as $\mathcal{O}(\min(nd^2,n^2d))$ and in memory storage as $\mathcal{O}(d^2 )$, where $d$ is the number of features (or basis functions if used) \cite{cline2006computation, golub2013matrix}.

\begin{figure}[!t]
\centering

\subfloat[]{\includegraphics[width=1.4in]{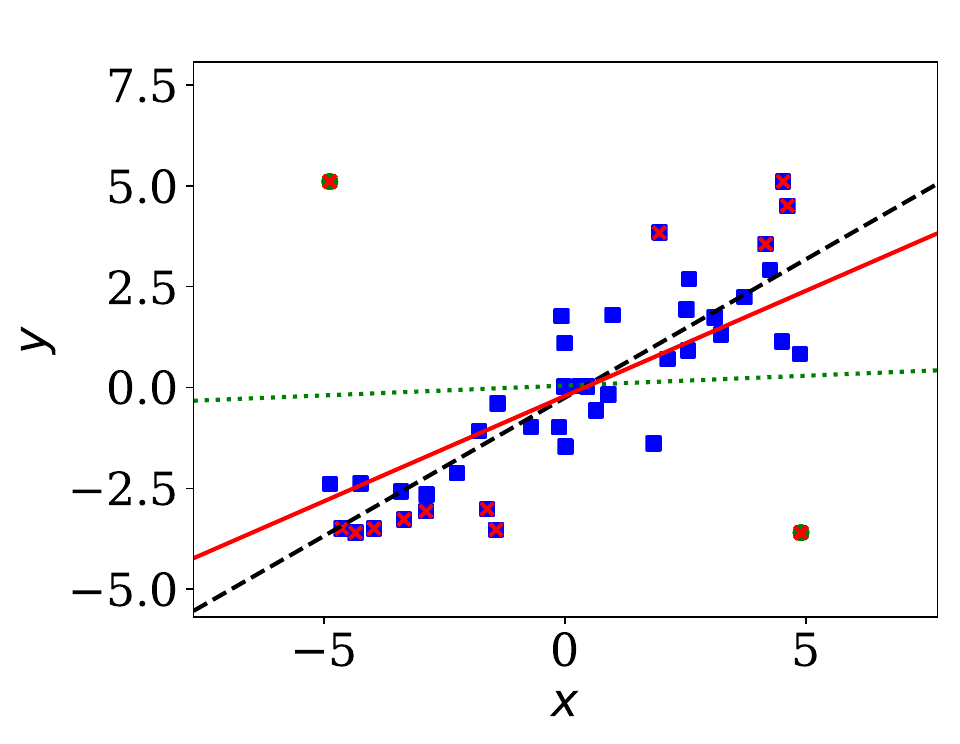}
}
\subfloat[]{\includegraphics[width=1.4in]{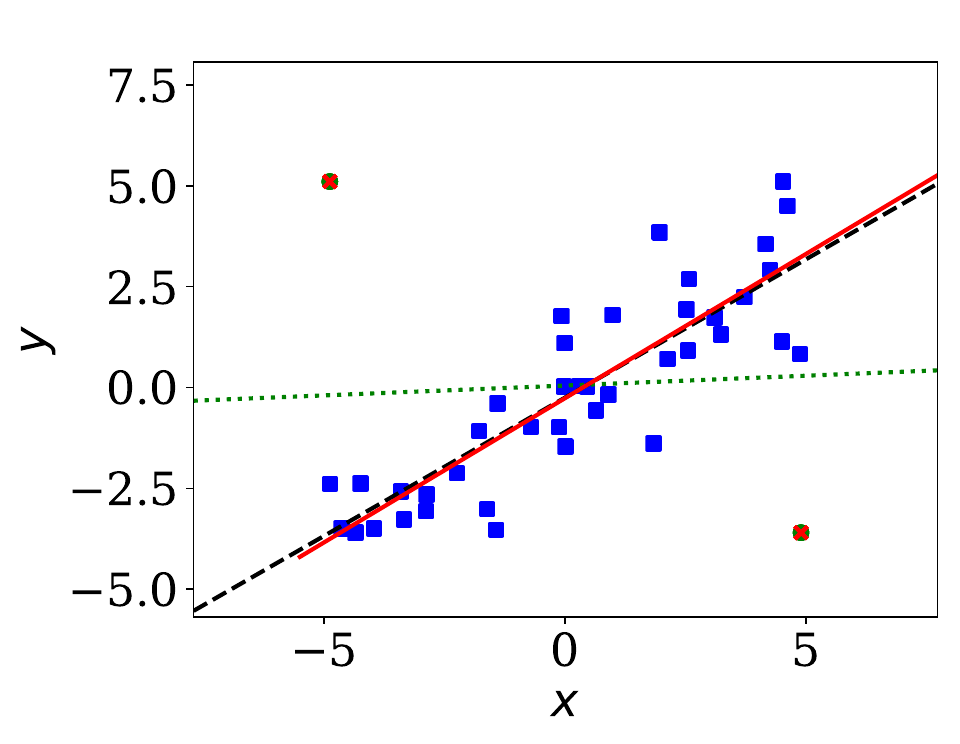}
}
\\
\vspace{-0.4cm}
\subfloat[]{\includegraphics[width=1.4in]{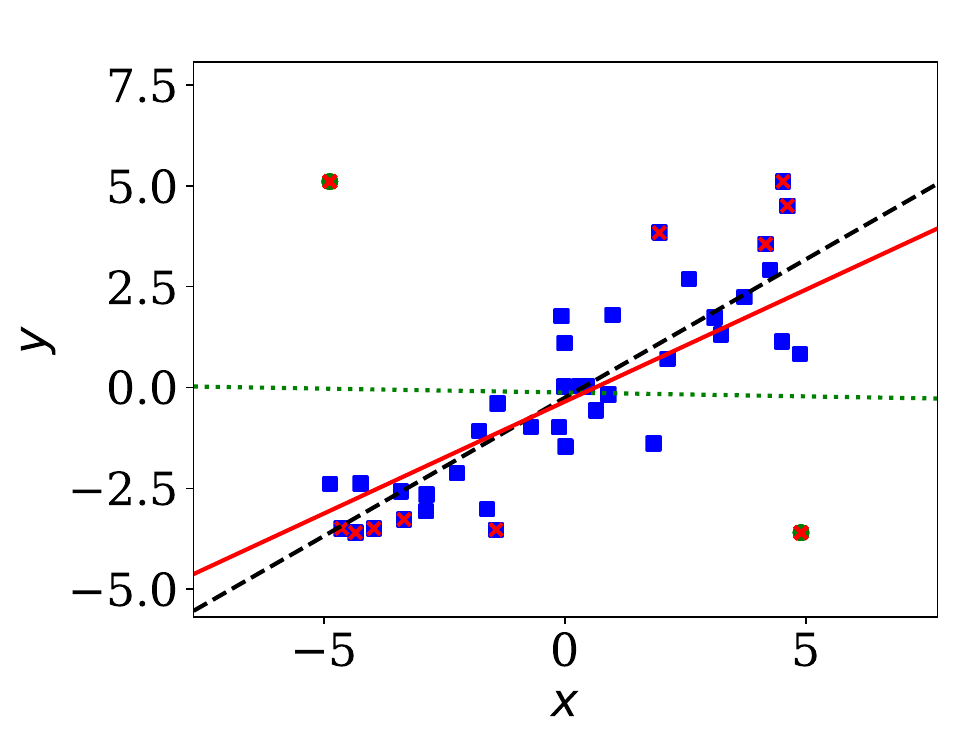}
}
\subfloat[]{\includegraphics[width=1.4in]{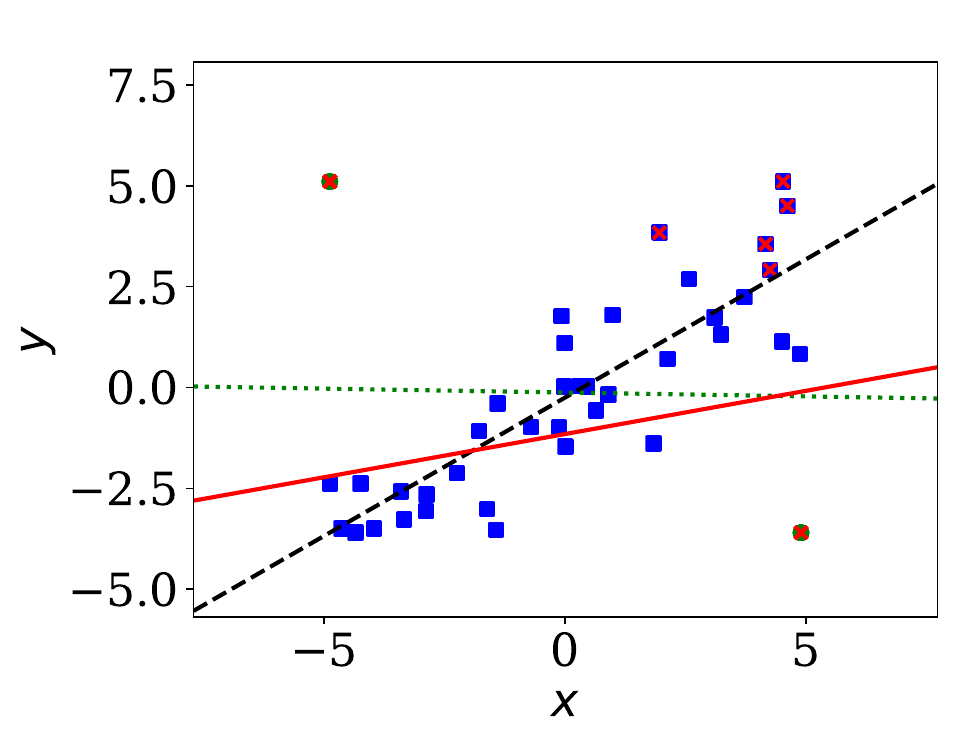}
}
\\
\vspace{-0.4cm}
\subfloat[]{\includegraphics[width=1.4in]{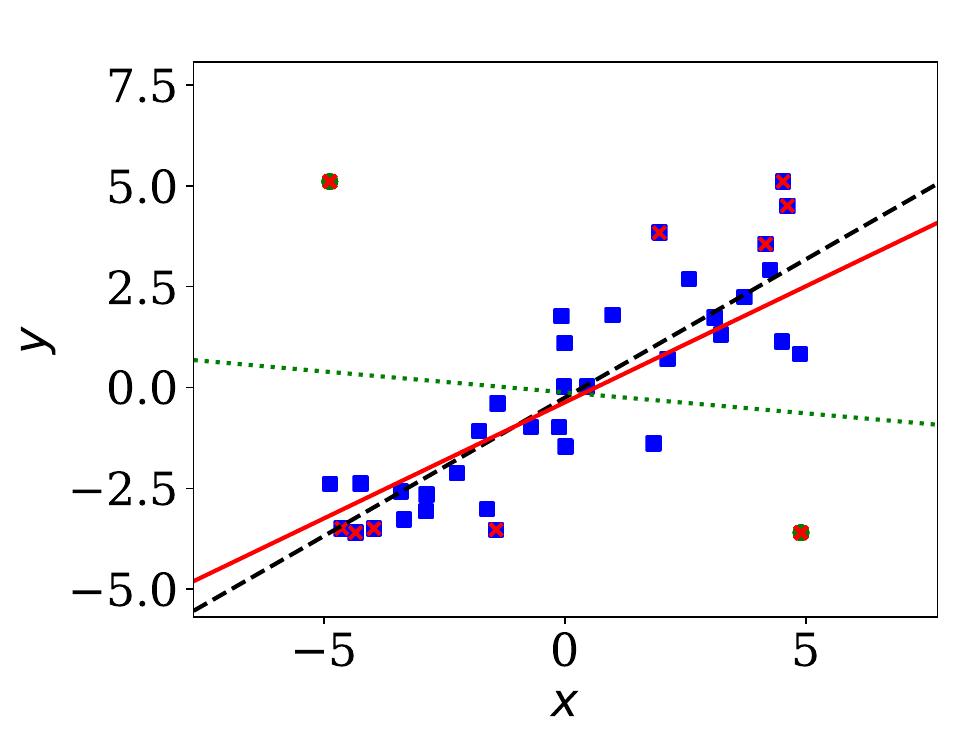}
}
\subfloat[]{\includegraphics[width=1.4in]{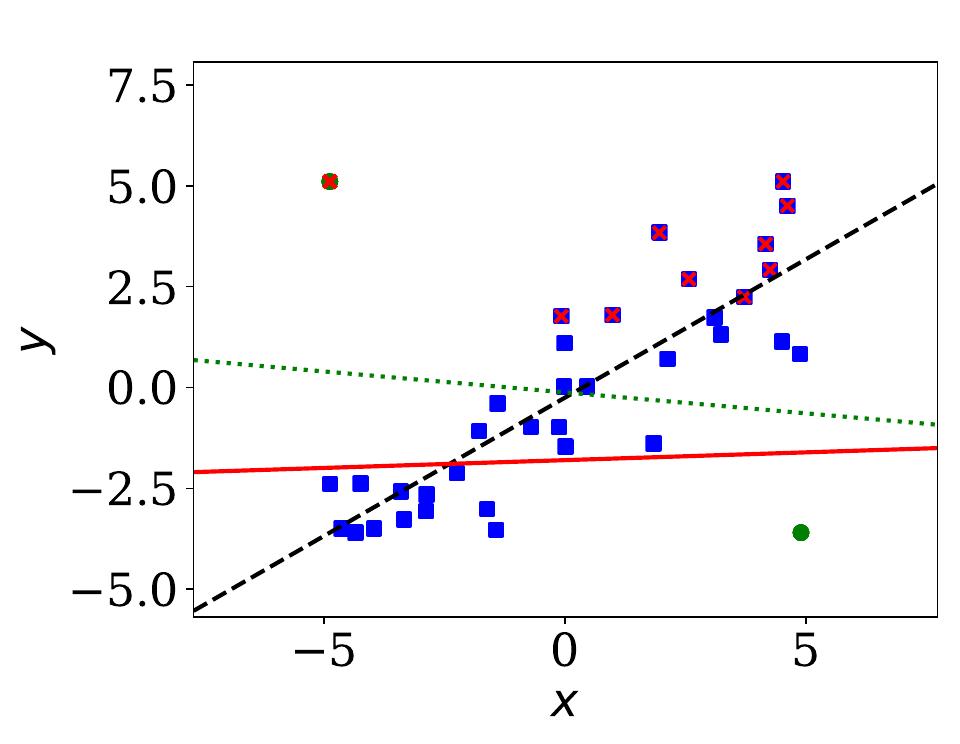}
}
\vspace{-.2cm}
\caption{Comparison of BayesClean and TRIM.
The blue points are clean points, green points are poisoning points, red crosses denote points rejected by the defense. The black dashed line is the regression learned from clean data, the green dotted line is the regression learned from poisoned data, and the red solid line is the regression learned under the training data not rejected by the defense. Varying defenses and the percentage of poisoning points: (a) BayesClean, $20\%$, (b) TRIM, $20\%$, (c) BayesClean, $24\%$  (d) TRIM, $24\%$, (e) BayesClean, $30\%$ and (f) TRIM, $30\%$. We can observe that BayesClean is more robust than TRIM for large ratios of poisoning points.
\vspace{-0.5cm}}
\label{fig:trim_bayesclean}
\end{figure}

In Fig.~\ref{fig:trim_bayesclean} we compare TRIM and BayesClean in a synthetic example, when the number of poisoning points is significant (i.e., $> 20\%$). For TRIM, we assume the defender knows the number of poisoning points, which, although unrealistic, favors it. Nevertheless, we observe there is a threshold (around $24\%$ poisoning points) above which TRIM fails to detect some poisoning points, even though it knows the number of poisoning points. In contrast, BayesClean is able to filter the poisoning points also for larger ratios of poisoning points.

TRIM, like other defenses, assumes that poisoning points have a larger error than the clean points on the poisoned model. So we believe TRIM fails because for large ratios of poisoning points, the mean error of the clean points approaches or even surpasses the mean error of the poisoning points on the poisoned model, as the model's parameters are shifted toward the poisoning points.
In contrast, the properties of the predictive variance of BayesClean makes it more robust than TRIM in such contexts. In the experimental evaluation below, we show that BayesClean performs better than state-of-the-art defenses when attacks consider detectability.

\subsection{Experimental Evaluation of BayesClean}

In this subsection we evaluate \emph{BayesClean}, our defense proposed, comparing it to state-of-the-art defenses: TRIM \cite{jagielski2018manipulating}, SEVER \cite{diakonikolas2019sever}, Proda \cite{wen2021great}, and  DUTI \cite{zhang2018training}. Note that, unlike other defenses, DUTI assumes the presence of a small clean, trusted set. We highlight this by representing it differently (with a dashed line) in the plots. See Appx.~\ref{subsec:expset2} for further details.

\subsubsection{Linear Regression}

\begin{figure*}[!t]
\centering
\subfloat[]{\includegraphics[width=1.55in]{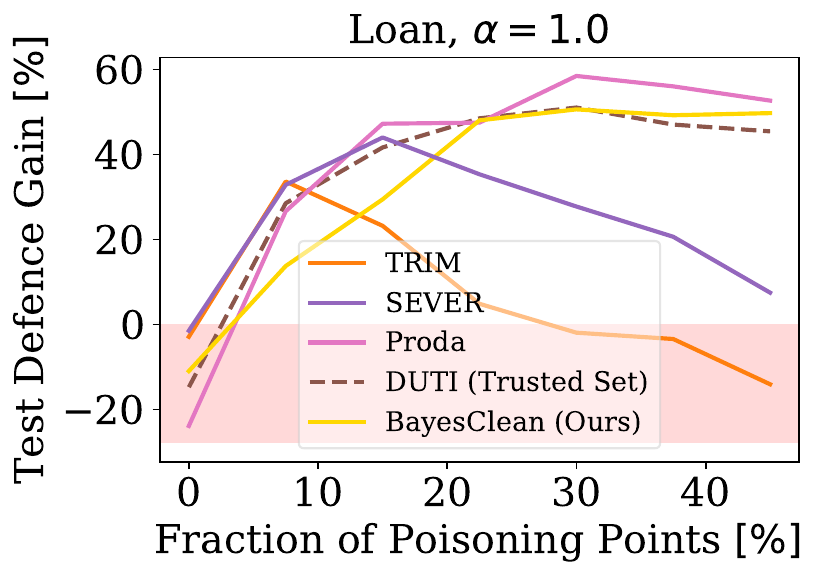}
\label{fig:mse_tst_def_a}}
\subfloat[]{\includegraphics[width=1.5in]{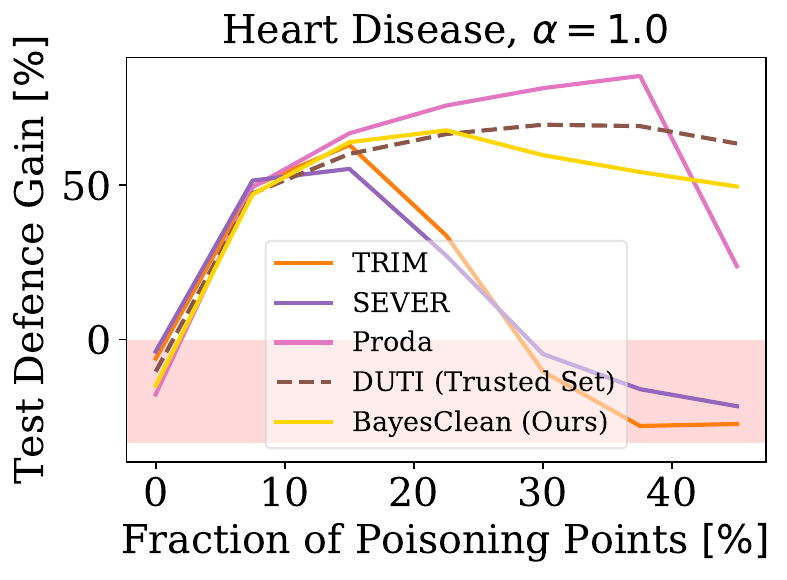}
\label{fig:mse_tst_def_b}}
\subfloat[]{\includegraphics[width=1.6in]{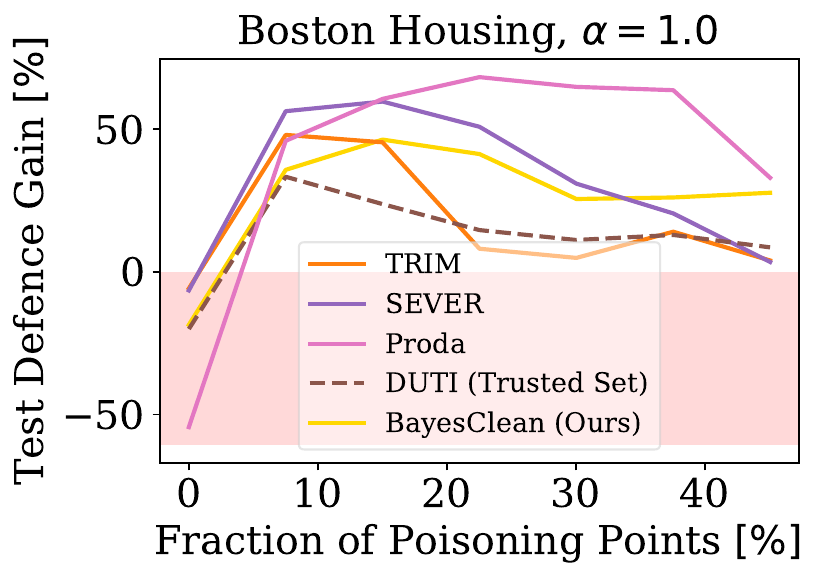}
\label{fig:mse_tst_def_c}}
\subfloat[]{\includegraphics[width=1.55in]{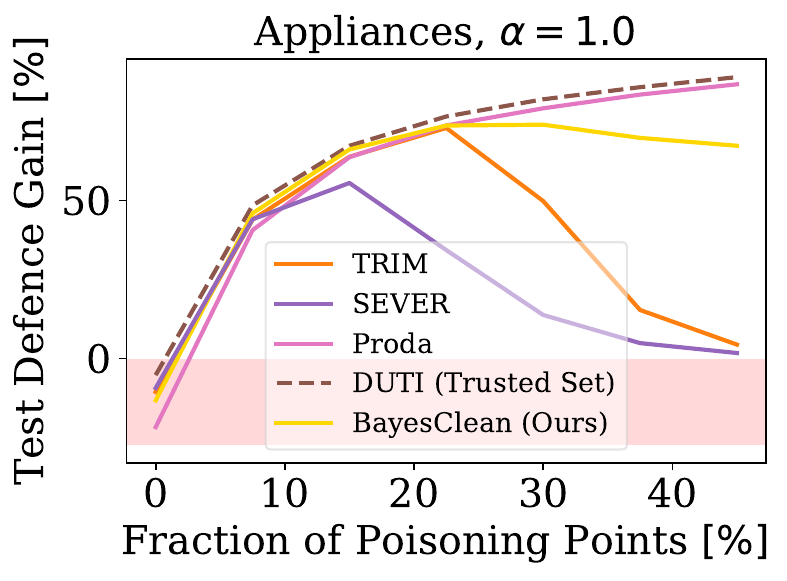}
\label{fig:mse_tst_def_d}}
\\
\vspace{-0.4cm}
\subfloat[]{\includegraphics[width=1.5in]{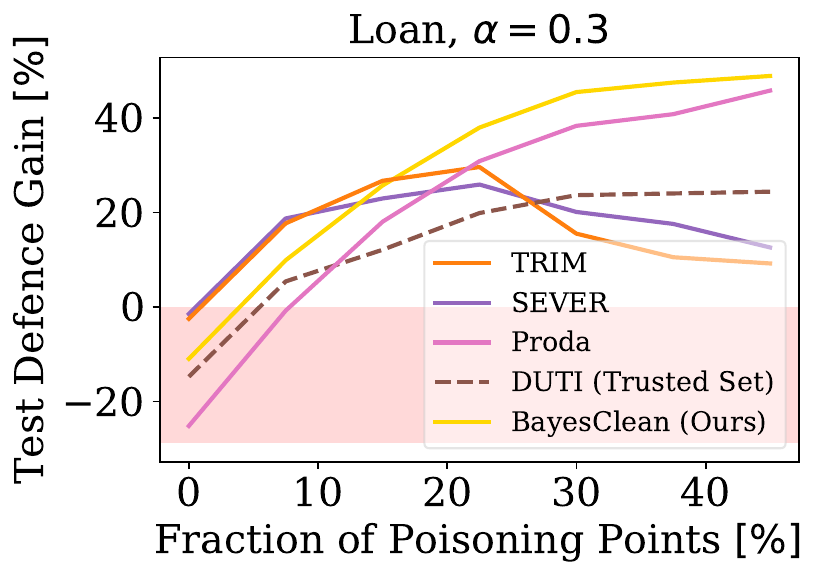}
\label{fig:mse_tst_def_e}}
\subfloat[]{\includegraphics[width=1.5in]{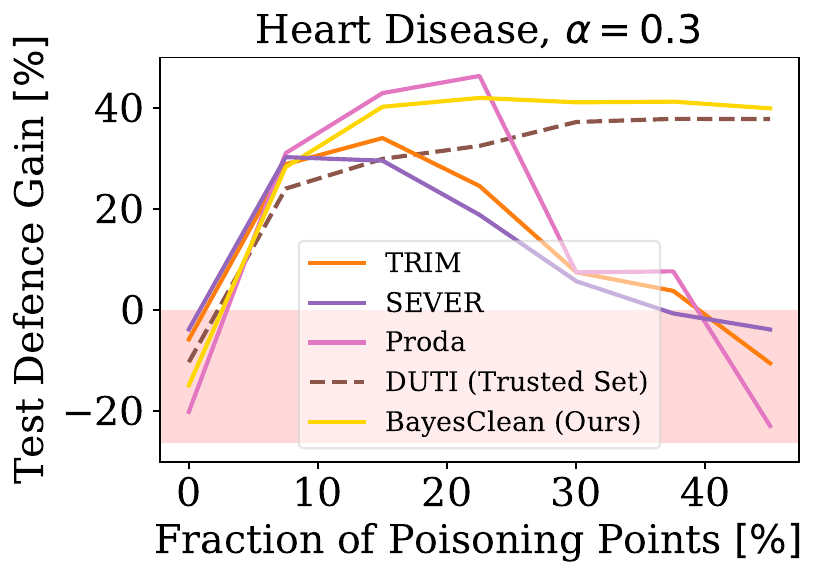}
\label{fig:mse_tst_def_f}}
\subfloat[]{\includegraphics[width=1.5in]{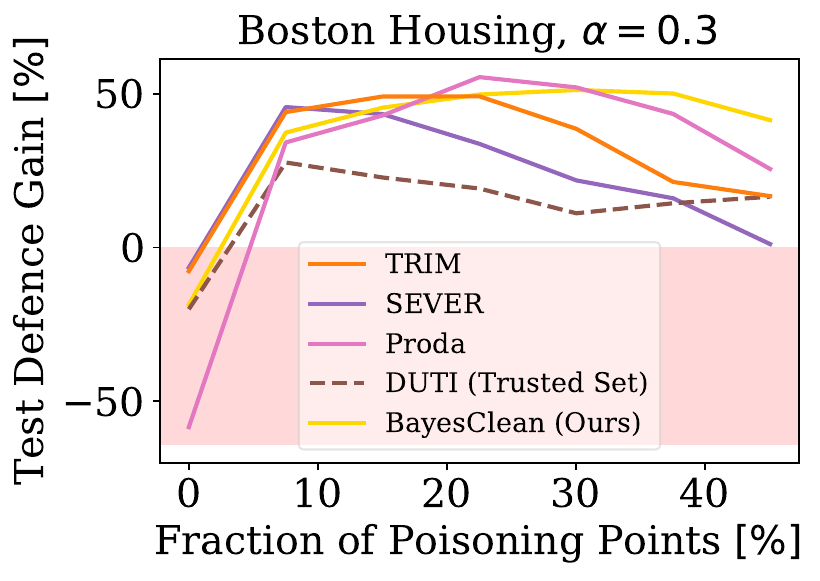}
\label{fig:mse_tst_def_g}}
\subfloat[]{\includegraphics[width=1.5in]{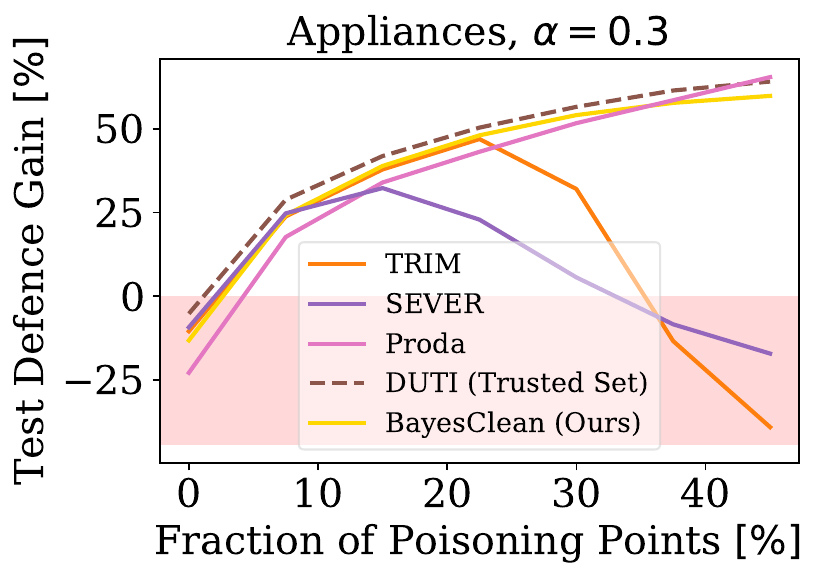}
\label{fig:mse_tst_def_h}}
\\
\vspace{-0.4cm}
\subfloat[]{\includegraphics[width=1.5in]{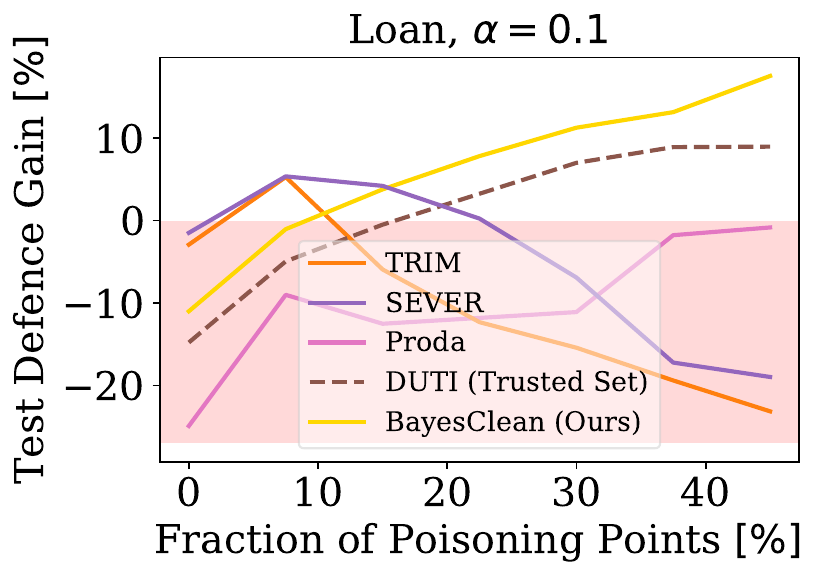}
\label{fig:mse_tst_def_i}}
\subfloat[]{\includegraphics[width=1.5in]{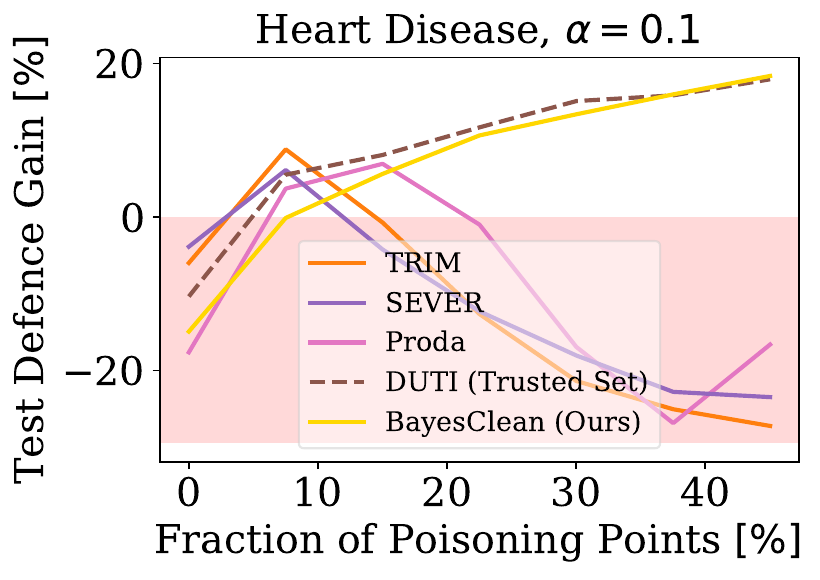}
\label{fig:mse_tst_def_j}}
\subfloat[]{\includegraphics[width=1.5in]{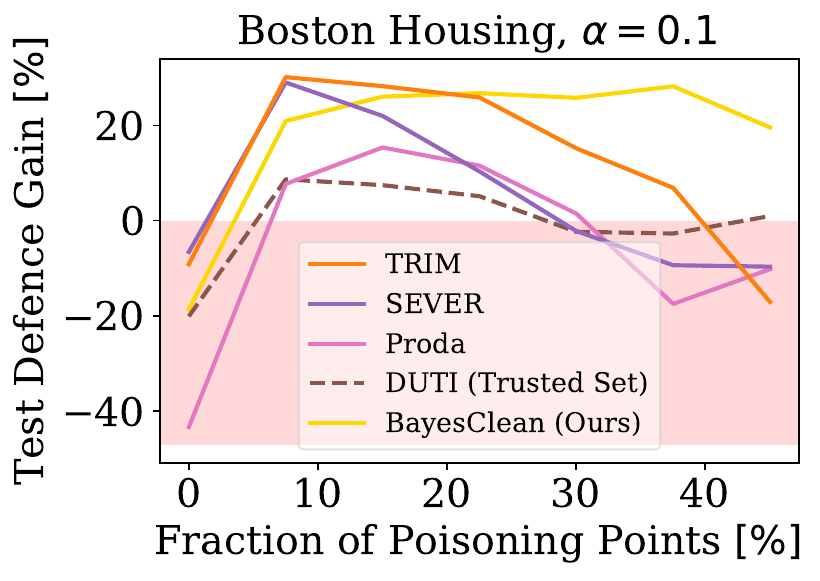}
\label{fig:mse_tst_def_k}}
\subfloat[]{\includegraphics[width=1.5in]{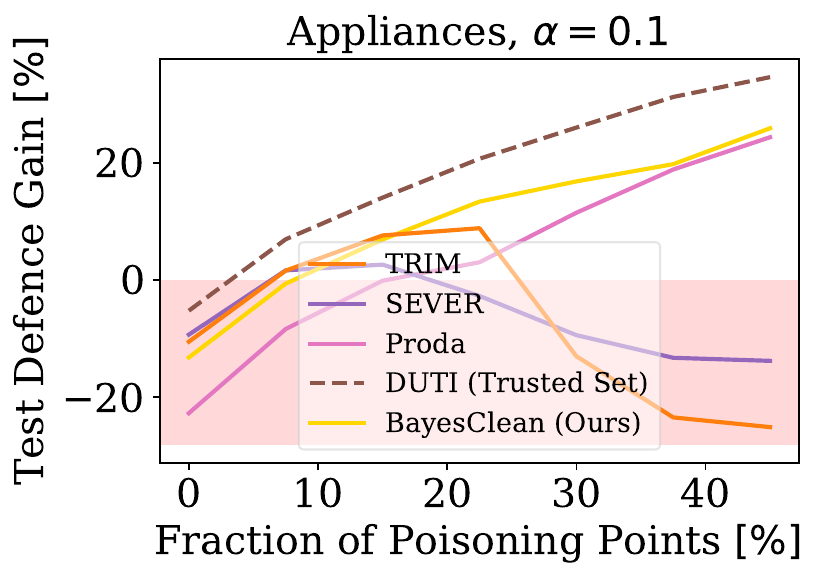}
\label{fig:mse_tst_def_l}}
\caption{Test defense gain of LR when using BayesClean, compared to state-of-the-art defenses. The first, second, third, and fourth columns correspond to Loan, Heart Disease, Boston Housing, and Appliances, respectively. The first, second, and third rows correspond to $\alpha = 1$, $\alpha = 0.3$, and $\alpha = 0.1$.
\vspace{-0.3cm}}
\label{fig:mse_tst_def}
\end{figure*}

Fig.~\ref{fig:mse_tst_def} show the performance of BayesClean compared to state-of-the-art defenses, when testing our attack against LR. In most cases, TRIM, SEVER, and Proda present a similar pattern: they perform well for low ratios of poisoning, and worse for large ratios. This corroborates our hypothesis that there is a threshold of poisoning points where the model is shifted too much towards the poisoning distribution to be useful to detect poisoning points. The error and/or the norm of the gradients of poisoning points can then be lower than error of the clean points on the poisoned model. We will explore this hypothesis further in future work.

On the other hand, the test defense gain of BayesClean is more stable and sometimes increases with the ratio of poisoning. That is, BayesClean is competitive with the other defenses for low ratios of poisoning and aggressive attacks ($\alpha=1$). Moreover, BayesClean improves upon most defenses when the number of poisoning points is large (e.g., larger than $20\%$ - $30\%$), even when the attacker considers detectability. DUTI also performs well because it is based on a trusted validation set, however it performs poorly on Boston Housing. These results show the benefits of our proposed defense and the importance of considering the uncertainty of the model in future research. As commented in $\S$~\ref{subsec:predvar}, the uncertainty (or variance) of the model can be more robust than the error of the model when the ratio of poisoning is large.

\subsubsection{Deep Neural Networks}

\begin{figure}[!t]
\centering
\subfloat[]{\includegraphics[width=1.45in]{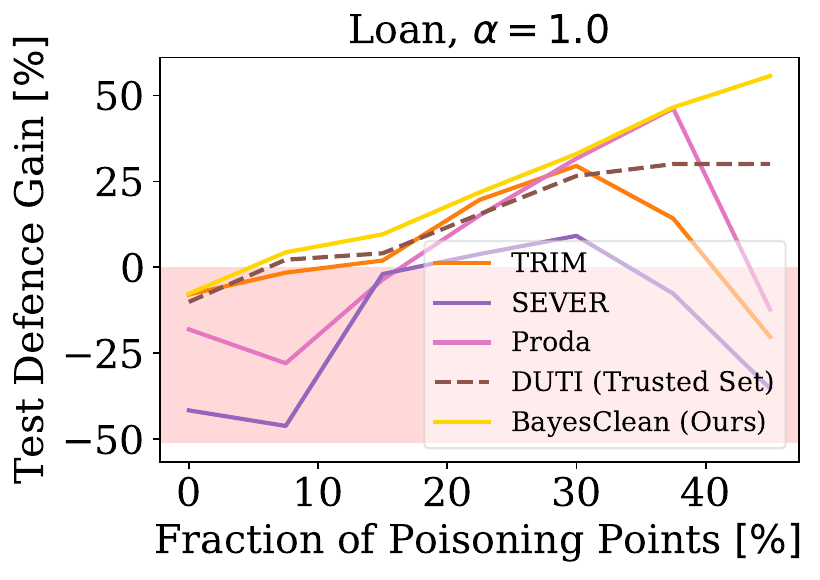}
\label{fig:mse_tst_dnn_def_a}}
\subfloat[]{\includegraphics[width=1.45in]{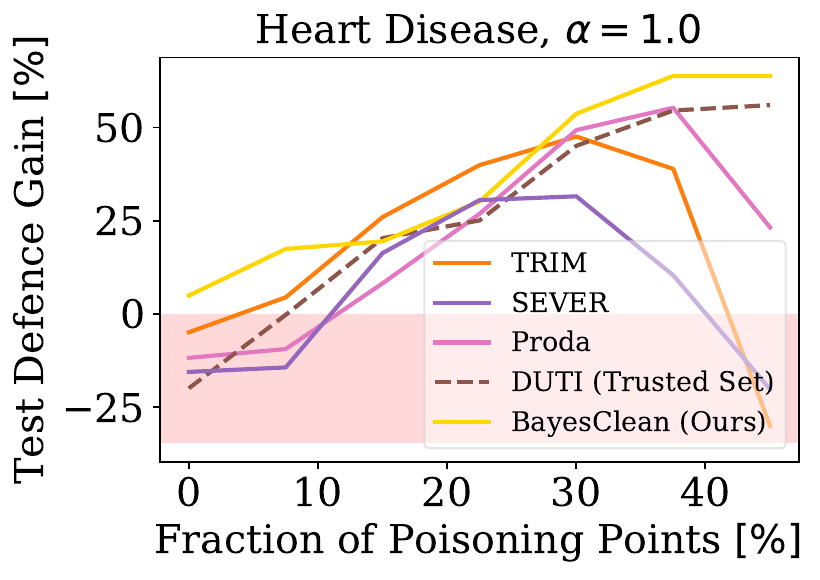}
\label{fig:mse_tst_def_dnn_d}}
\\
\vspace{-0.4cm}
\subfloat[]{\includegraphics[width=1.45in]{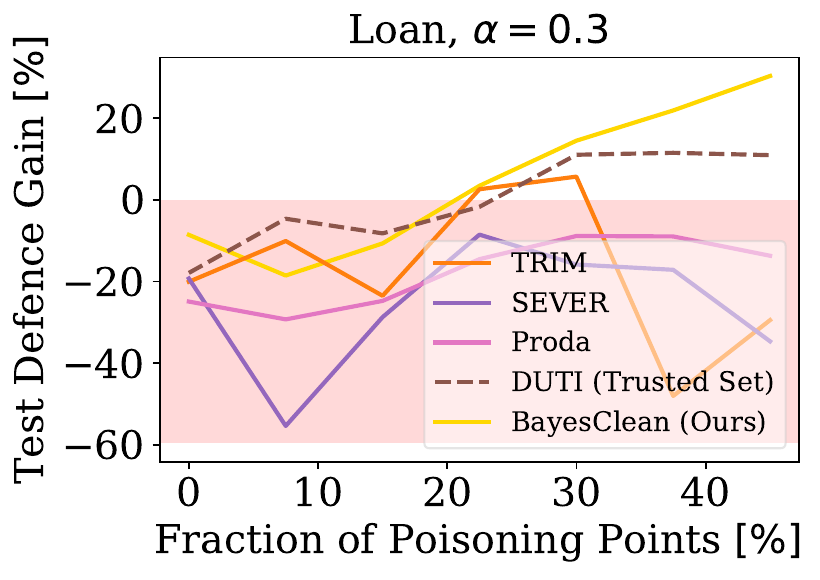}
\label{fig:mse_tst_dnn_def_b}}
\subfloat[]{\includegraphics[width=1.45in]{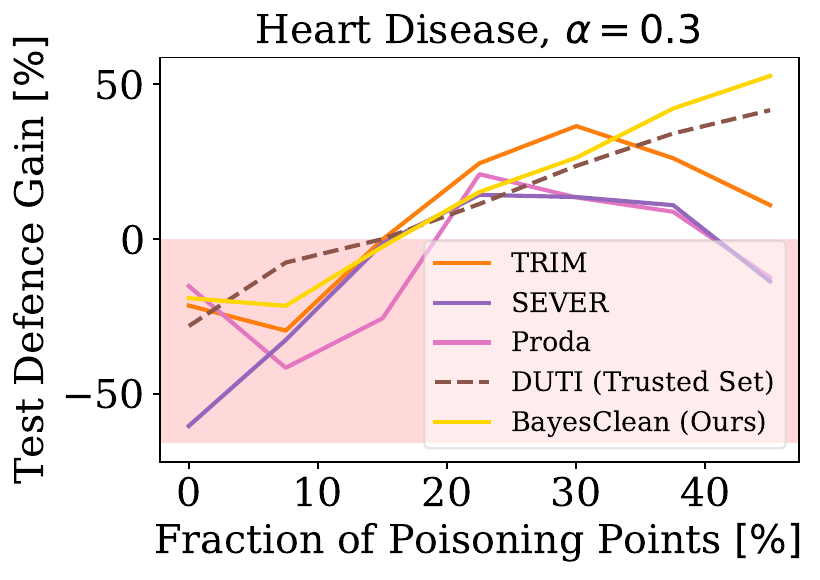}
\label{fig:mse_tst_dnn_def_e}}
\\
\vspace{-0.4cm}
\subfloat[]{\includegraphics[width=1.45in]{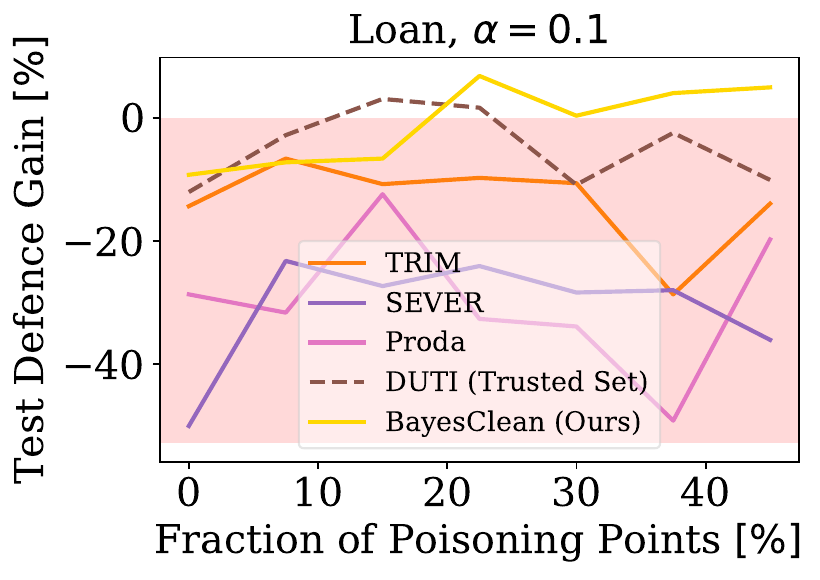}
\label{fig:mse_tst_dnn_def_c}}
\subfloat[]{\includegraphics[width=1.45in]{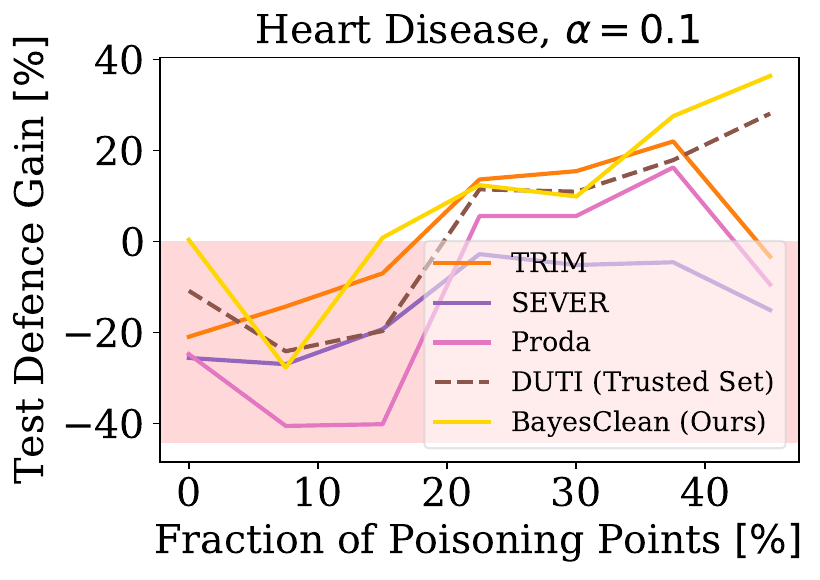}
\label{fig:mse_tst_dnn_def_f}}
\vspace{-.2cm}
\caption{Testing defense gain of the DNNs when using BayesClean, compared to state-of-the-art defenses. The first column shows the results for Loan, and the second shows the results for Heart Disease. The first, second, and third rows show the results for $\alpha=1$, $\alpha=0.3$, and $\alpha=0.1$, correspondingly. \vspace{-0.3cm}}
\label{fig:mse_tst_dnn_def}
\end{figure}

For completeness, in Fig.~\ref{fig:mse_tst_dnn_def} we show the performance of our defense on DNNs. Even though \emph{BayesClean} is based on a Bayesian LR model, its predictive variance is able to detect poisoning points targeting DNNs. As in the case of LR, our defense is competitive in most cases, being particularly useful when the attacker considers detectability and the ratio of poisoning points is large. It is also evident, according to the negative values of the test defense gain, that defending DNNs is more challenging. These results suggest further research is needed on methods to defend DNNs in regression settings.

\section{Conclusions}

Most previous work in regression poisoning has focused on unconstrained and non-adaptive attacks that just aim to maximize the error of the model. These attacks can be mitigated by deploying defense mechanisms that consider the distribution of the training data as shown in the related work. However, such attacks are not realistic as the attacker also aims to remain undetected.
The vulnerability of regression models to such attacks was unknown.

To cover this research gap, we have proposed a novel attack formulation based on a multiobjective bilevel optimization problem. This problem models the two conflicting objectives of the attacker: maximizing attack effectiveness and minimizing the risk of being detected. This formulation is more realistic and is parameterizable allowing to investigate different attacker trade-offs. However, solving this attack formulation is challenging. We have introduced new heuristics, optimizations and a novel normalization process to address this challenge.

In particular, we have proposed an effective method to solve the optimization problem, by using RMD,
and a novel approach to the normalization of objectives. Our experiments in real-world datasets show that stealthy attacks may be less effective but they bypass proposed defenses and are sufficient to induce significant errors in the models. Defenses are in some cases not worth investing in as the resulting model has poorer performance than when no defense is employed. Hence, the deployment of defenses may induce a false sense of security. Most defenses fail against stealthy attacks, but some also fail in the context in which they are claimed to work, e.g., for large ratios of poisoning (e.g., more than 25\% in some cases).

Finally, we have proposed a novel defense against poisoning attacks in regression settings called BayesClean. Fundamentally, this defense builds upon a Bayesian LR model whose hyperparameters are learned from the training set. Then, points are rejected depending on their relative location within the predictive variance of the model. Our experiments show the effectiveness of the proposed defense and its advantage w.r.t. other defenses when the ratio of poisoning points is significant, even when the attacker considers detectability.

Despite their importance in many practical problems, we are still in the early stages of understanding the attack surface of regression models to poisoning attacks and developing robust techniques for their defense. Numerous issues need to be investigated and challenges to be addressed. We expect that the work presented here, will enable more realistic and systematic testing of the robustness to poisoning of ML models and the development of new and more robust defenses.

\bibliographystyle{IEEEtran}
\bibliography{refs}

{\appendices

\section{Normalization of Objective Functions}
\label{sec:norm_obj}

Alg.~\ref{alg:norm_obj} and Alg.~\ref{alg:pois2} describe the procedure of normalizing the objective functions introduced in $\S$~\ref{subsubsec:norm_obj}. In Alg.~\ref{alg:norm_obj}, 
$T_\text{out}$ refers to the number of iterations for the projected hypergradient ascent algorithm that computes the poisoning points, and $ \mathcal{P}$ denotes the set of indices of the clean training points to be replaced by the poisoning points. 
The functionality of Alg.~\ref{alg:pois2} 
is similar to Alg.~\ref{alg:norm_obj}.

\begin{algorithm*}[t]
	\caption{Computation of $B_\mathrm{p}$ Poisoning Points ($\alpha = 1$) and of the Normalization Coefficient for the Detectability-Risk Function}
	\label{alg:norm_obj}

    \begin{flushleft}
	{\bfseries Input:} $\mathcal{M}$, $\mathcal{A}$, $\mathcal{R}$, $\mathcal{L}$, $\mathcal{D}_\mathrm{val}$, $\mathcal{D}_\mathrm{tr}$,  $n_\mathrm{p}$, $B_\mathrm{p}$, $\{ \mathcal{P}^{(i)} \}_{i=1}^{n_{B_\mathrm{p}}}$, $T_\mathrm{out}$, $T$, $\gamma$, $\eta$ \\
		{\bfseries Output:} $ \mathcal{R}_\mathrm{ref}$
	\end{flushleft}
	
	\begin{algorithmic}[1]
	
	        \State  $n_{B_\mathrm{p}} \leftarrow \lceil {n_\mathrm{p}/B_\mathrm{p}}\rceil $
	
				\State $\alpha \leftarrow 1$ \Comment{Set maximum effectiveness and minimum detectability for the reference}
				\State $\mathcal{A}_{\mathrm{ref}_{\alpha=1}}  \leftarrow 1$
					\State $\mathcal{R}_{\mathrm{ref}_{\alpha=1}} \leftarrow 1$

		\For{$i=0$ \textbf{to} $n_{B_\mathrm{p}}-1$}

  \State 	${\bf w}^{(0)} \leftarrow \texttt{initW}(\mathcal{M})$ \Comment{Initialize {\bf w}}
		
		\For{$t=0$ \textbf{to} $T-1$}
		
		\State ${\bf g} \leftarrow  \nabla_{{\bf w}}\mathcal{L}({\bf w}^{(t)})$ 
		
		\State ${\bf w}^{(t+1)}\leftarrow {\bf w}^{(t)} - \eta {\bf g}$ \Comment{Stochastic Gradient Descent}
		\EndFor
		
			\State  $\mathcal{D}_\mathrm{p}^{(i)} \leftarrow \texttt{projHypGradAsc}(		\mathcal{M}, \mathcal{L}, \mathcal{L}_{\mathrm{ref}_{\alpha=1}}, \mathcal{R}, \mathcal{R}_{\mathrm{ref}_{\alpha=1}}, \mathcal{D}_\mathrm{val},
   \mathcal{D}_\mathrm{tr},  B_\mathrm{p}, \mathcal{P}^{(i)}, T_\mathrm{out}, T, \alpha, \gamma, \eta)$
	
	\State $\mathcal{R}_\mathrm{ref}^{(i)}  \leftarrow \mathcal{R}(\mathcal{D}_\mathrm{p}^{(i)})$
	
				\State $\mathcal{D}_\mathrm{tr}  \leftarrow ( \mathcal{D}_\mathrm{tr} \setminus  \{({\bf x}_{\mathrm{tr}_k} ,y_{\mathrm{tr}_k})\}_{k\in \mathcal{P}^{(i)}} ) \cup \mathcal{D}_\mathrm{p}^{(i)}$ \Comment{Replace $B_\mathrm{p}$ samples of $\mathcal{D}_\mathrm{\mathrm{tr}}$ by  $\mathcal{D}_\mathrm{p}^{(i)}$ }
		
		\EndFor
		
		\State $\mathcal{R}_\mathrm{ref} \leftarrow \max_i  \{ \mathcal{R}_\mathrm{ref}^{(i)} \}_{i=1}^{n_{B_\mathrm{p}}}$
	
	\end{algorithmic}
	
\end{algorithm*}

\begin{algorithm*}[t]
	\caption{Computation of $B_\mathrm{p}$ Poisoning Points ($\alpha\neq 1$) and of the Normalization Coefficient for the Effectiveness Function}
	\label{alg:pois2}

    \begin{flushleft}
	{\bfseries Input:} $\mathcal{M}$, $\mathcal{L}$, $\mathcal{R}$, $ \mathcal{R}_{\mathrm{ref}}$  $\mathcal{D}_\mathrm{val}$, $\mathcal{D}_\mathrm{tr}$,  $n_\mathrm{p}$, $B_\mathrm{p}$, $\{ \mathcal{P}^{(i)} \}_{i=1}^{n_{B_\mathrm{p}}}$, $T_\mathrm{out}$, $T$, $\alpha$, $\gamma$, $\eta$ \\
		{\bfseries Output:} $ \mathcal{D}_\mathrm{tr}$
	\end{flushleft}
	
	\begin{algorithmic}[1]
	
	        \State  $n_{B_\mathrm{p}} \leftarrow \lceil {n_\mathrm{p}/B_\mathrm{p}}\rceil $

		\For{$i=0$ \textbf{to} $n_{B_\mathrm{p}}-1$}
		
				\State 	${\bf w}^{(0)} \leftarrow \texttt{initW}(\mathcal{M})$ \Comment{Initialize {\bf w}}
		
		\For{$t=0$ \textbf{to} $T-1$}
		
		\State ${\bf g} \leftarrow  \nabla_{{\bf w}}\mathcal{L}({\bf w}^{(t)})$ 
		
		\State ${\bf w}^{(t+1)}\leftarrow {\bf w}^{(t)} - \eta {\bf g}$ \Comment{Stochastic Gradient Descent}
		\EndFor
		
						\State $\mathcal{L}_\mathrm{ref}^{(i)}  \leftarrow \mathcal{L}(\mathcal{D}_\mathrm{val},    {\bf w}^{(T)})$

\State  $\mathcal{D}_\mathrm{p}^{(i)} \leftarrow \texttt{projHypGradAsc}(		\mathcal{M}, \mathcal{L}, \mathcal{L}_{\mathrm{ref}}^{(i)}, \mathcal{R}, \mathcal{R}_{\mathrm{ref}},  \mathcal{D}_\mathrm{val}, \mathcal{D}_\mathrm{tr},  B_\mathrm{p}, 
\mathcal{P}^{(i)}, T_\mathrm{out}, T, \alpha, \gamma, \eta)$

			\State $\mathcal{D}_\mathrm{tr}  \leftarrow ( \mathcal{D}_\mathrm{tr} \setminus  \{({\bf x}_{\mathrm{tr}_k} ,y_{\mathrm{tr}_k})\}_{k\in \mathcal{P}^{(i)}} ) \cup \mathcal{D}_\mathrm{p}^{(i)}$ \Comment{Replace $B_\mathrm{p}$ samples of $\mathcal{D}_\mathrm{\mathrm{tr}}$ by  $\mathcal{D}_\mathrm{p}^{(i)}$ }
	
		\EndFor

	\end{algorithmic}
	
\end{algorithm*}

\section{Reverse-Mode Differentiation for Multiobjective Optimization in Regression Poisoning}

\label{sec:rmd_mult}

Alg.~\ref{alg:bg2} details the RMD procedure used to calculate the hypergradients of the attacker's objective function ($\mathcal{A}_d$) w.r.t. the poisoning points. Alg.~\ref{alg:bg2} requires first to train the learning algorithm for $T$ training iterations. Then, the hypergradients estimate is computed by differentiating the updates of the learning algorithm and reversing its sequence of parameters. We use a notation similar to \cite{domke2012generic, maclaurin2015gradient, munoz2017towards, carnerero2024hyperparameter}, where more details on the derivation of this algorithm can be found.

\begin{algorithm*}[!t]
	\caption{Reverse-Mode Differentiation for Multiobjective Optimization in Regression Poisoning}
	\label{alg:bg2}
	
	\begin{flushleft}
	
    {\bfseries Input:}  $\mathcal{M}$, $\mathcal{A}$, $\mathcal{A}_\mathrm{ref}$, $\mathcal{R}$, $\mathcal{R}_\mathrm{ref}$, $\mathcal{L}$,  $\mathcal{D}_\mathrm{val}$, $\mathcal{D}_\mathrm{tr}$,  $\mathcal{D}_\mathrm{p}^{(\tau)}$, ${\bf w}^{(0)}$,  $T$, $\eta$ \\
		{\bfseries Output:} $\nabla_{{\bf X}_\mathrm{p}}\mathcal{A}_d({\bf w}^{(T)})$, $\nabla_{{\bf y}_\mathrm{p}}\mathcal{A}_d({\bf w}^{(T)})$
		
	\end{flushleft}
	
	\begin{algorithmic}[1]
		
		\For{$t=0$ \textbf{to} $T-1$}
		
		\State ${\bf g} \leftarrow  \nabla_{{\bf w}}\mathcal{L}({\bf w}^{(t)})$ 
		
		\State ${\bf w}^{(t+1)}\leftarrow {\bf w}^{(t)} - \eta {\bf g}$ \Comment{Stochastic Gradient Descent}
		\EndFor
		
		\State $\mathcal{A}_\mathrm{norm}({\bf w}^{(T)}) \leftarrow \mathcal{A}({\bf w}^{(T)})/\mathcal{A}_\mathrm{ref}$
		
				\State $\mathcal{R}_\mathrm{norm}({\bf w}^{(T)}) \leftarrow \mathcal{R}({\bf w}^{(T)})/\mathcal{R}_\mathrm{ref}$
		
        \State $\mathcal{A}_{d_\mathrm{norm}}({\bf w}^{(T)})  \leftarrow \alpha \mathcal{A}_\mathrm{norm}({\bf w}^{(T)}) - (1 - \alpha) \mathcal{R}_\mathrm{norm}({\bf w}^{(T)})$ 
        
		\State $d{\bf X}_\mathrm{p}^{(T)} \leftarrow \nabla_{{\bf X}_\mathrm{p}}\mathcal{A}_{d_\mathrm{norm}}({\bf w}^{(T)})$
		
		\State $d{\bf y}_\mathrm{p}^{(T)} \leftarrow \nabla_{{\bf y}_\mathrm{p}}\mathcal{A}_{d_\mathrm{norm}}({\bf w}^{(T)})$
		
		\State $d{\bf w}^{(T)} \leftarrow \nabla_{{\bf w}}\mathcal{A}_{d_\mathrm{norm}}({\bf w}^{(T)})$
		\For{$t=T-1$ \textbf{down to} $0$}
		
		\State $d{\bf g}_{\mathrm{w}} \gets \left(\nabla^2_{{\bf w}}\mathcal{L}({\bf w}^{(t)})\right)d{\bf w}^{(t+1)} $ \Comment{Hessian-vector product}
		\State $d{\bf g}_{\mathrm{X}_\mathrm{p}} \gets \left(\nabla_{{\bf X}_\mathrm{p}}\nabla_{{\bf w}}\mathcal{L}({\bf w}^{(t)})\right)^{\textsf{T}} d{\bf w}^{(t+1)} $ \Comment{Hessian-vector product}
		
				\State $d{\bf g}_{{\text y}_\mathrm{p}} \gets \left(\nabla_{{\bf y}_\mathrm{p}}\nabla_{{\bf w}}\mathcal{L}({\bf w}^{(t)})\right)^{\textsf{T}} d{\bf w}^{(t+1)}_\mathrm{p} $ \Comment{Hessian-vector product}
		
				\State	$
		d{\bf w}^{(t)}  \leftarrow d{\bf w}^{(t+1)}  - \eta d{\bf g}_{\mathrm{w}}
		$
		
		\State	$
		d{\bf X}_\mathrm{p}^{(t)}  \leftarrow d{\bf X}_\mathrm{p}^{(t+1)}  - \eta d{\bf g}_{\mathrm{X}_\mathrm{p}}
		$
		
				\State	$
		d{\bf y}_\mathrm{p}^{(t)}  \leftarrow d{\bf y}_\mathrm{p}^{(t+1)}  - \eta d{\bf g}_{{\text y}_\mathrm{p}}
		$
		
		\EndFor
		\State $ \nabla_{{\bf X}_\mathrm{p}}\mathcal{A}_{d_\mathrm{norm}}({\bf w}^{(T)})
		\leftarrow d{\bf X}_\mathrm{p}^{(0)}$
		
				\State $ \nabla_{{\bf y}_\mathrm{p}}\mathcal{A}_{d_\mathrm{norm}}({\bf w}^{(T)})
		\leftarrow d{\bf y}_\mathrm{p}^{(0)}$

	\end{algorithmic}
\end{algorithm*}

\section{Projected Hypergradient Ascent for Regression Poisoning}
\label{sec:proj}

Alg.~\ref{alg:hyperreg_} describes the procedure to solve the multiobjective bilevel problem (\ref{eqAttacker_}) proposed in the paper. Essentially, this algorithm implements projected hypergradient descent/ascent for $T_\text{out}$ iterations to optimize, in a coordinated manner, the poisoning points---replaced into the training set.

To reduce the computational burden, we consider the simultaneous optimization of a batch of $n_\text{p}$ poisoning points, $\mathcal{D}_\text{p} = \{({\bf x}_{\text{p}_k} ,y_{\text{p}_k})\}^{n_\text{p}}_{k=1}$. We generate the initial values of $\mathcal{D}_\text{p}$ by cloning $n_\text{p}$ samples---uniformly sampled without duplicates---of $\mathcal{D}_\text{tr}$. This process is carried out in the function \texttt{initDp}.
Then, these $n_\text{p}$ poisoning samples replace the $n_\text{p}$ clean samples of $\mathcal{D}_\text{tr}$ whose indices are in the set $\mathcal{P}$.

To solve the bilevel problem, every time the variables in the outer problem are updated, the model's parameters need to be previously initialized and optimized. Thus, let \texttt{initW} 
be a particular initialization for the model's parameters. $\texttt{hypGrad}$ refers to the particular optimization algorithm used to train the model's parameters and compute the corresponding hypergradients. In this work, this algorithm is RMD (Alg.~\ref{alg:bg2}).

\begin{algorithm*}[t]
	\caption{Projected Hypergradient Ascent for Regression Poisoning}
	\label{alg:hyperreg_}
	
    \begin{flushleft}
	{\bfseries Input:} $\mathcal{M}$, $\mathcal{A}$, $\mathcal{A}_\mathrm{ref}$, $\mathcal{R}$, $\mathcal{R}_\mathrm{ref}$, $\mathcal{L}$, $\mathcal{D}_\mathrm{val}$, $\mathcal{D}_\mathrm{tr}$,  $B_\mathrm{p}$, $\mathcal{P}$,  $T_\mathrm{out}$, $T$, $\alpha$, $\gamma$, $\eta$ \\
		{\bfseries Output:} $\mathcal{D}_\mathrm{p}^{(T_\mathrm{out})}$
	\end{flushleft}
	
	\begin{algorithmic}[1]
	
	\State $\mathcal{D}_\mathrm{p}^{(0)} \leftarrow \texttt{initDp}(\mathcal{D}_\mathrm{tr}, B_\mathrm{p})$ \Comment{Generate $B_\mathrm{p}$ initial samples for $\mathcal{D}_\mathrm{p}^{(0)}$ }

			\State $\mathcal{D}_\mathrm{tr}^{(0)}  \leftarrow ( \mathcal{D}_\mathrm{tr} \setminus  \{({\bf x}_{\mathrm{tr}_k} ,y_{\mathrm{tr}_k})\}_{k\in \mathcal{P}} ) \cup \mathcal{D}_\mathrm{p}^{(0)}$ \Comment{Replace $B_\mathrm{p}$ samples of $\mathcal{D}_\mathrm{\mathrm{tr}}$ by  $\mathcal{D}_\mathrm{p}^{(0)}$ }

		\label{lin:initl_}

		\For{$\tau=0$ \textbf{to} $T_\mathrm{out}-1$}
		
		\State 	${\bf w}^{(0)} \leftarrow \texttt{initW}(\mathcal{M})$ \Comment{Initialize {\bf w}}
		\label{lin:initw1_}
		\State $ \nabla_{{\bf X}_\mathrm{p}}\mathcal{A}_{d_\mathrm{norm}}({\bf w}^{(T)}), \nabla_{{\bf y}_\mathrm{p}}\mathcal{A}_{d_\mathrm{norm}}({\bf w}^{(T)})
		\leftarrow \texttt{hypGrad}(\mathcal{M},\mathcal{L}, \mathcal{L}_\mathrm{ref}, \mathcal{R}, \mathcal{R}_\mathrm{ref}, { \mathcal{D}_\mathrm{val}, \mathcal{D}_\mathrm{tr}, \mathcal{D}_\mathrm{p}^{(\tau)}}, {\bf w}^{(0)}, T, \eta)$
		\label{lin:hypgrad_} \Comment{RMD (Alg.~\ref{alg:bg2}) 
		}	
		\State ${\bf X}_\mathrm{p}^{(\tau+1)}\leftarrow \Pi_{\Phi (\mathcal{D}_\mathrm{p})} \left({\bf X}_\mathrm{p}^{(\tau)} + \gamma\nabla_{{\bf X}_\mathrm{p}}\mathcal{A}_{d_\mathrm{norm}}({\bf w}^{(T)})\right)$ \Comment{Projected Hypergradient Ascent}
		\label{lin:phga_}
		
				\State ${\bf y}_\mathrm{p}^{(\tau+1)}\leftarrow \Pi_{\Phi (\mathcal{D}_\mathrm{p})} \left({\bf y}_\mathrm{p}^{(\tau)} + \gamma\nabla_{{\bf y}_\mathrm{p}}\mathcal{A}_{d_\mathrm{norm}}({\bf w}^{(T)})\right)$ \Comment{Projected Hypergradient Ascent}
		\label{lin:phga2_}

		\State $\mathcal{D}_\mathrm{tr}^{(\tau+1)}  \leftarrow ( \mathcal{D}_\mathrm{tr}^{(\tau)} \setminus \mathcal{D}_\mathrm{p}^{(\tau)} ) \cup \mathcal{D}_\mathrm{p}^{(\tau+1)}$ \Comment{Update $\mathcal{D}_\mathrm{\mathrm{tr}}^{(\tau)}$ with $\mathcal{D}_\mathrm{p}^{(\tau+1)}$ }
		
		\label{lin:upddtr2_}
	
		\EndFor
	
	\end{algorithmic}
	
\end{algorithm*}

\section{BayesClean Algorithm}

\label{sec:bayesclean_alg}

Alg.~\ref{alg:bayesclean} shows the algorithm for BayesClean, the novel defense proposed in $\S$~\ref{sec:bayesclean}. In Alg.~\ref{alg:bayesclean}, $\text{diag}({\bf X}{\boldsymbol \Sigma}_\text{post} {\bf X}^{\textsf{T}})$ denotes the vector of entries of the main diagonal of ${\bf X}{\boldsymbol \Sigma}_\text{post} {\bf X}^{\textsf{T}}$.

\begin{algorithm*}[t]
	\caption{BayesClean}
	\label{alg:bayesclean}
	
    \begin{flushleft}
	{\bfseries Input:} $\mathcal{D}=({\bf X},{\bf y})$, $\lambda_1= \lambda_2=\beta_1=\beta_2=10^{-6}$, $c_1$, $c_2, T_\text{EM}$   \\
		{\bfseries Output:} $\mathcal{I}_1$, $\mathcal{I}_2$, $\mathcal{I}_3$
	\end{flushleft}
	
	\begin{algorithmic}[1]
	
	\State $\hat{\lambda}, \hat{\beta} \in \argmax_{\lambda, \beta}  \log p({\bf y}|{\bf X}, \lambda, \beta) + \log p(\lambda)  + \log p(\beta)$ \Comment{Learn hyperparameters of the Bayesian LR model via EM, up to $T_\text{EM}$ iterations}

 \State  ${\boldsymbol \Sigma}_{\text{post}} \leftarrow (\hat{\beta} {\bf X}^{\textsf{T}} {\bf X} + \hat{\lambda} {\bf I}_d)^{-1}, \ {\boldsymbol \mu}_{\text{post}} \leftarrow \hat{\beta} {\boldsymbol \Sigma}_{\text{post}} {\bf X}^{\textsf{T}} {\bf y}$ \Comment{Update posterior distribution}

 \State ${\boldsymbol \mu}_* \leftarrow  {\bf X}{\boldsymbol \mu}_\text{post}, \ {\boldsymbol \sigma}_*^2 \leftarrow \hat{\beta}^{-1} + \text{diag}({\bf X}{\boldsymbol \Sigma}_\text{post} {\bf X}^{\textsf{T}})$ \Comment{Compute posterior predictive distribution for each input point}
 \State $\mathcal{I}_1 \leftarrow \{i \ | \ |y_i| \leq |\mu_{*_i} + c_1\sigma_{*_i}|\} $  \Comment{Indices of points to accept}
  \State $\mathcal{I}_2 \leftarrow \{i \ | \  |\mu_{*_i} + c_1\sigma_{*_i}| < |y_i| \leq |\mu_{*_i} + c_2\sigma_{*_i}|\} $  \Comment{Indices of points to flag}
   \State $\mathcal{I}_3 \leftarrow \{i \ | \ |y_i| > |\mu_{*_i} + c_2\sigma_{*_i}|\} $  \Comment{Indices of points to reject}	
	\end{algorithmic}
 \vspace{-0.1cm}
\end{algorithm*}

\section{Experimental Settings}

\label{subsec:expset2}

All our results are the average of $10$ repetitions with different random data splits for training, validation and test sets. For each repetition, we normalize the datasets w.r.t. the mean and standard deviation of the clean training data.\footnote{The analysis of attacks when the normalization process is also poisoned is left as future work.} For all the attacks, we measure the average test NMSE for different attack strengths, where the number of poisoning points ranges from $0\%$ to $45\%$.  The size of the batch of poisoning points that are simultaneously optimized is included in Table~\ref{tabAttack2}. 
In this way, we simulate seven different ratios of poisoning ranging from $0\%$ to $45\%$.

We simulate different ratios of poisoning points in a cumulative manner: Once the optimization of the current batch of poisoning points and hyperparameters is finished,\footnote{The criterion to finish the loop that optimises the variables of the outer level problem is given by the number of hyperiterations.} this batch of poisoning points is fixed and the next batch of poisoning points is replaced into the remaining clean training set, whereas the hyperparameters are re-initialized, to carry out their corresponding optimization.  To accelerate their optimization, the hypergradients for the poisoning points are normalized with respect to their $L_2$ norm.

The  parameters of the LR models are always initialized with zeros, for all the datasets. The DNN models have two hidden layers with Leaky ReLU activation functions as follows: $(\text{\# Features})\times32\times8\times1$. 
In the DNN models, these parameters are initially filled with values according to Xavier Initialization method \cite{glorot2010understanding}, using a uniform distribution for all the parameters except the bias terms, which are initialized with a value of $10^{-2}$. 

\begin{table*}[t]
	\centering	
		\begin{tabular}{|l|c|c|c|c|}
			\hline
			Dataset & \makecell{\# Training Samples} & \makecell{\# Validation Samples}  & \makecell{\# Test Samples} & \# Features\\
			\hline
			Loan & $3,361$ & $1,008$ & $2,353$ & $88$ \\
			Heart Disease & $1,599$ & $480$ & $1,119$ & $28$ \\
			Boston Housing & $253$ & $76$ & $177$ & $13$  \\
   			Appliances & $9,868$ & $2,960$ & $6,907$ & $16$ \\
			\hline
	\end{tabular} 
        \vspace{.2cm}
  	\caption{Datasets used in the experiments}
  	\label{tabDatasets2}
\end{table*}

\begin{table*}[!t]	
	\centering	
		\begin{tabular}{|l|c|c|c|c|c|}
  \hline
			Dataset (Model) &  $T_\mathrm{out}$ &  $\gamma$  & $T$  & $\eta$  & Poisoning Batch \\
			\hline
			Loan (LR) & $120$ & $0.9$ &  $30$ & $0.1$ & $252$ \\
			Heart Disease (LR) & $50$ & $0.9$  & $40$ & $0.1$ & $120$ \\
			Boston Housing (LR) & $100$  & $0.9$  & $40$ & $0.1$ & $19$ \\
   			Appliances (LR) & $50$  & $0.9$ & $70$ & $0.2$ & $740$ \\
			Loan (DNN) & $150$ & $0.9$ & $90$ & $0.1$ & $252$\\
			Heart Disease (DNN) & $80$ & $0.9$ & $100$  &  $0.1$ & $120$ \\
			\hline
	\end{tabular}
        \vspace{.2cm}
	\caption{Experimental settings for crafting the poisoning attack.}
 \label{tabAttack2}
\end{table*}

\begin{table*}[!t]
	\centering
		\begin{tabular}{|l|c|c|}
			\hline
			Dataset (Model) &  $\eta_\mathrm{tr}$ &  {\#} Epochs \\
			\hline
			Loan (LR) & $0.1$ & $30$ \\
			Heart Disease (LR) & $0.1$ & $40$ \\
			Boston Housing (LR) & $0.1$ & $40$ \\
			Appliances (LR) & $0.2$ & $70$ \\
				Loan (DNN) & $0.1$ & $90$ \\
			Heart Disease (DNN) & $0.1$ &  $100$  \\
			\hline 
	\end{tabular} 
    \vspace{.2cm}
 	\caption{Experimental settings for testing the attacks.}
  	\label{tabTrain2}
\end{table*}

For all the experiments we make use of SGD to both update the parameters in the forward pass of RMD, and train the model when testing the attack (full batch training). The choice of the number of iterations for the inner problem, $T$, depends on the model and the training dataset. 
The details of the attack settings are shown in Table~\ref{tabAttack2}, whereas the ones for testing the attacks are in Table~\ref{tabTrain2}. 

With respect to $L_2$ regularization, 
we tested a methodology akin to \cite{jagielski2018manipulating} and \cite{friedman2010regularization}, where the authors use $K$-fold cross-validation to select the value of $\lambda$, and the clean data is used both for training and validation in an unbiased way. However, in our experiments in the datasets mentioned before, the  value of the regularization hyperparameter learned following this approach was very low, so the performance was very similar to the one obtain without using regularization. Therefore, in the following experiments we set $\lambda=0$.

In the synthetic examples, the features of the clean points are generated according to a uniform distribution $\mathcal{U}(-5, 5)$, the labels are generated according to the transformation: $y=0.8x$, and the additive noise of the labels follows a Gaussian distribution $\mathcal{N}(0.0, 1.2^2)$. The LR model and TRIM are trained using SGD with a learning rate $\eta=0.1$ and $\# \text{epochs} = 40$ (full batch).

We evaluate our stealthy attack against several benchmark defenses: TRIM \cite{jagielski2018manipulating}, Huber \cite{huber1964robust}, and SEVER \cite{diakonikolas2019sever}. Then, we compare our novel defense, BayesClean, with  TRIM \cite{jagielski2018manipulating}, SEVER \cite{diakonikolas2019sever}, Proda \cite{wen2021great}, and DUTI \cite{zhang2018training}. One of the drawbacks of these defenses is that they require to define the number of points to reject beforehand. Since we are testing attacks corresponding to up to $45\%$ of poisoning points, we consider strong defenses that reject $40\%$ of the input points. We also recall that DUTI 
is guided through a small trusted clean set, which represents an advantage w.r.t. other defenses.

We set the hyperparameters of the defenses as follows: for BayesClean, we set $\lambda_1=\lambda_2=\beta_1=\beta_2=10^{-6}$, $c_1=0.5$, $c_2=1.5$, and $T_\text{EM}=300$ (maximum number of iterations when learning the hyperparameters). For TRIM, SEVER, and Proda, we use the same training hyperparameters as in Table~\ref{tabTrain2}. For Proda, we set $\epsilon_\text{Proda}=10^{-5}$ (upper bound on the probability that no group of points belongs to the clean training set) and $\gamma_\text{Proda}=5$ (size of each group of points), considering its time complexity. For DUTI, we set $\lambda_\text{DUTI}=0.1$ (regularization hyperparameter), $c_\text{DUTI}=100$ (confidence level on all trusted items) and the number of trusted items---sampled from the validation set---as for $101$ for Loan, $48$ for Heart Disease, $8$ for Boston Housing, and $296$ for Appliances (ratio w.r.t. the training set size similar to \cite{zhang2018training}). For Huber, we set $\lambda_\text{Huber}=10^{-4}$ (regularization hyperparameter), $T_\text{Huber}=10^{4}$ (maximum number of iterations), and we explore values for $\epsilon_\text{Huber}$ (hyperparameter that controls the number of samples that should be classified as outliers) in the range $[1.1, 10]$.

All the experiments have been run on $2 \times 11$~GB NVIDIA GeForce\textregistered \hspace{0cm}  GTX 1080 Ti GPUs. The RAM memory is $64$~GB ($4\times16$~GB) Corsair VENGEANCE DDR4 $3000~\mathrm{MHz}$. The processor (CPU) is Intel\textregistered \hspace{0cm} Core\texttrademark \hspace{0cm} i7 Quad Core Processor i7-7700k ($4.2$~GHz) $8$~MB Cache.

\section{Additional Results}

\label{sec:addit_results}

\subsection{Stealthy Attacks Against Linear Regression}

\label{subsec:addit_results_lr}

\begin{figure}[!t]
\centering
\subfloat[]{\includegraphics[width=1.7in]{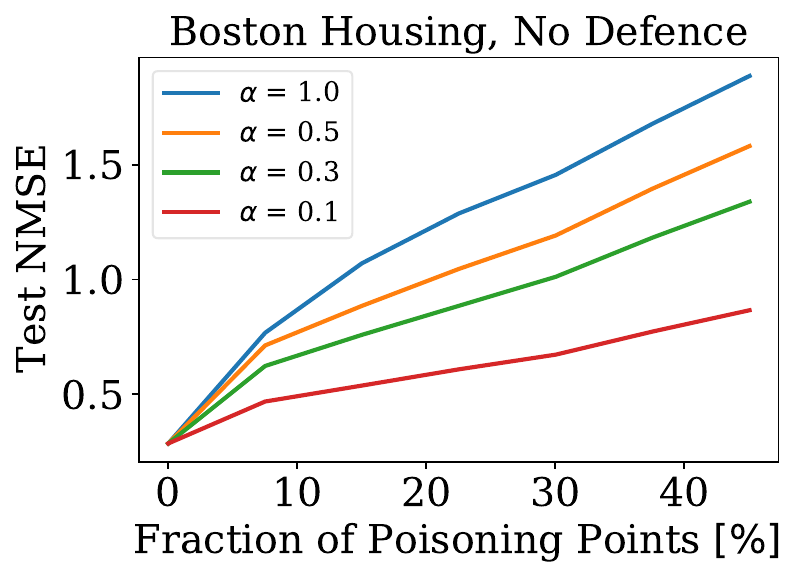}
\label{fig:mse_tst_alphas_d}}
\subfloat[]{\includegraphics[width=1.6in]{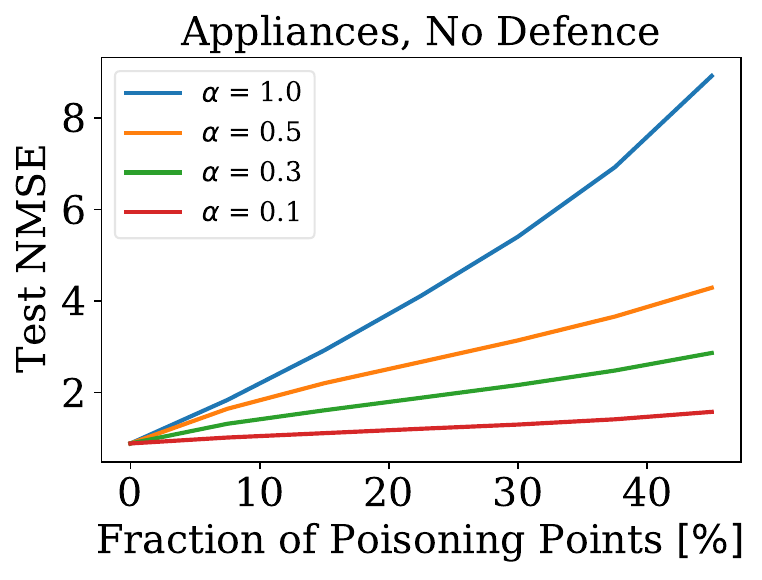}
\label{fig:mse_tst_alphas_e}}
\vspace{-.2cm}
\caption{Test NMSE of LR when there is no defense deployed, for $\alpha=1$,  $\alpha=0.5$,  $\alpha=0.3$, and  $\alpha=0.1$. (a) Boston Housing. (b) Appliances.
\vspace{-0.2cm}}
\label{fig:mse_tst_alphas2}
\end{figure}

Fig.~\ref{fig:mse_tst_alphas2} 
show the results 
for the stealthy attacks against LR corresponding to the \emph{base case} where defenses have not been deployed. These results are similar to the ones in Fig.~\ref{fig:mse_tst_alphas}. We observe that the test NMSE increases with the value of $\alpha$, as we generate more points that are out of distribution.

\subsection{Stealthy Attacks Against Deep Neural Networks}

\label{subsec:stealth_dnn_}

Fig.~\ref{fig:mse_tst_dnn} 
show the results 
for the stealthy attacks against DNNs corresponding to the \emph{base case} where defenses have not been deployed.
The results are similar to the ones in Fig.~\ref{fig:mse_tst_alphas} and Fig.~\ref{fig:mse_tst_alphas2}: the test NMSE increases with $\alpha$, which provides flexibility to generate a smooth transition between inliers and outliers. Interestingly, the test NMSE is larger compared to LR, suggesting that the DNN is more vulnerable to the poisoning attack.

\begin{figure}[!t]
\centering
\subfloat[]{\includegraphics[width=1.7in]{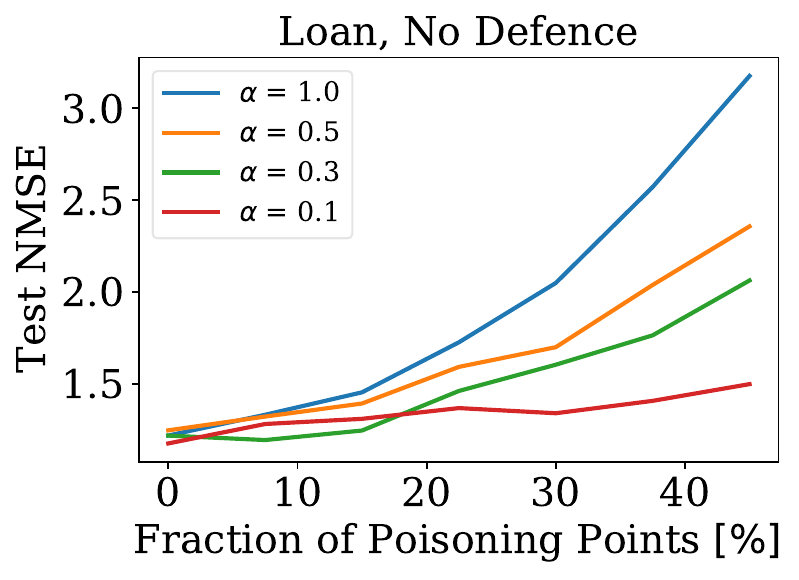}
\label{fig:mse_tst_dnn_a}}
\subfloat[]{\includegraphics[width=1.6in]{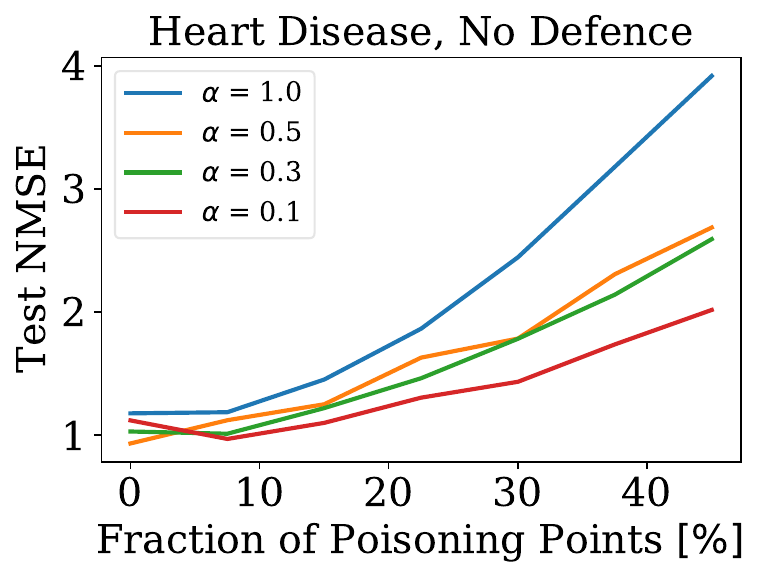}
\label{fig:mse_tst_dnn_b}}
\vspace{-.2cm}
\caption{Test NMSE of the DNNs when there is no defense deployed, for $\alpha=1$,  $\alpha=0.5$,  $\alpha=0.3$, and  $\alpha=0.1$.  (a) Loan. (b) Heart Disease.\vspace{-.3cm}}
\label{fig:mse_tst_dnn}
\end{figure}

\subsection{Predictive Variance under Data Poisoning}

\label{subsec:predvar_}

Fig.~\ref{fig:bayesclean} 
shows the intuition of BayesClean, and how the predictive variance increases with the number of poisoning points injected in the training set, as explained in $\S$~\ref{subsec:predvar}. In Fig.~\ref{fig:bayesclean}, the purple region corresponds to the condition $\mathcal{I}_1$ and the red region corresponds to
$\mathcal{I}_2$. 

\begin{figure}[!t]
\centering
\subfloat[]{\includegraphics[width=1.5in]{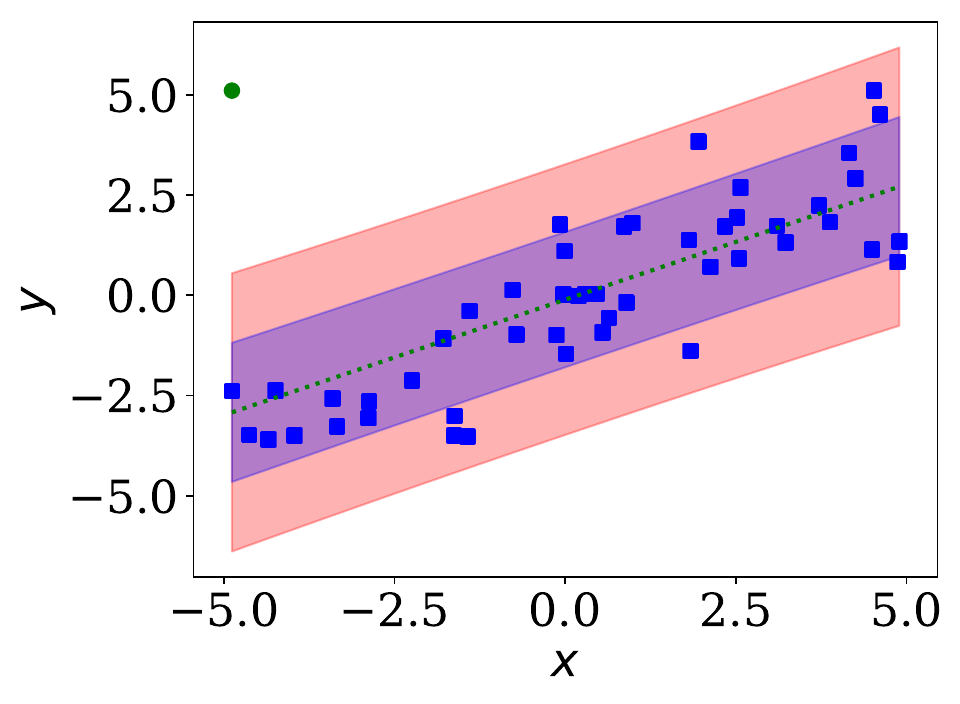}}
\\
\vspace{-0.4cm}
\subfloat[]{\includegraphics[width=1.5in]{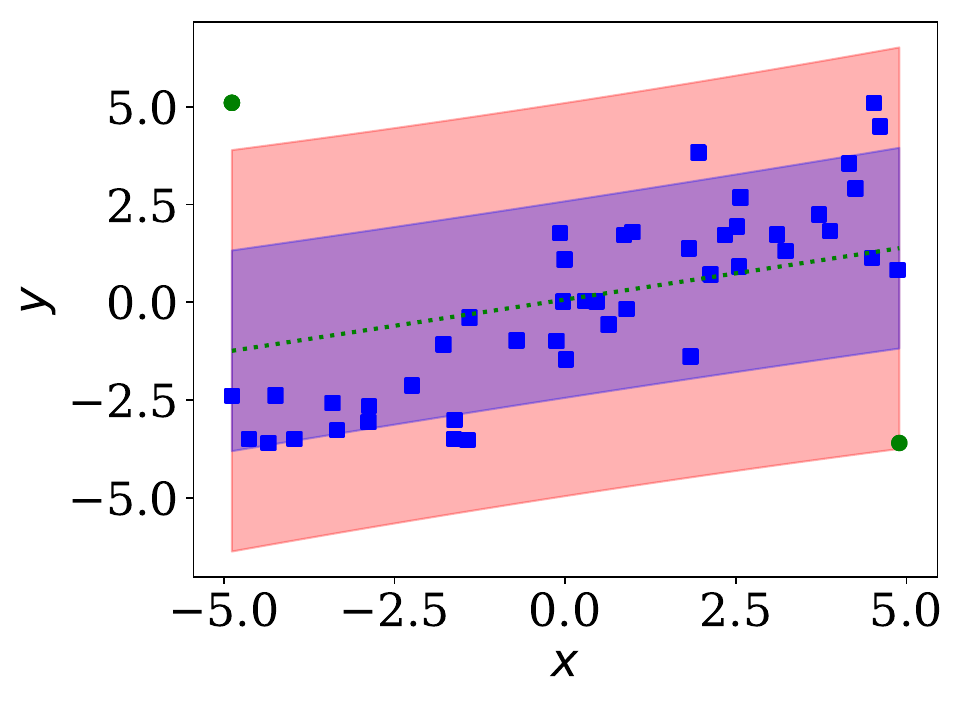}}
\subfloat[]{\includegraphics[width=1.5in]{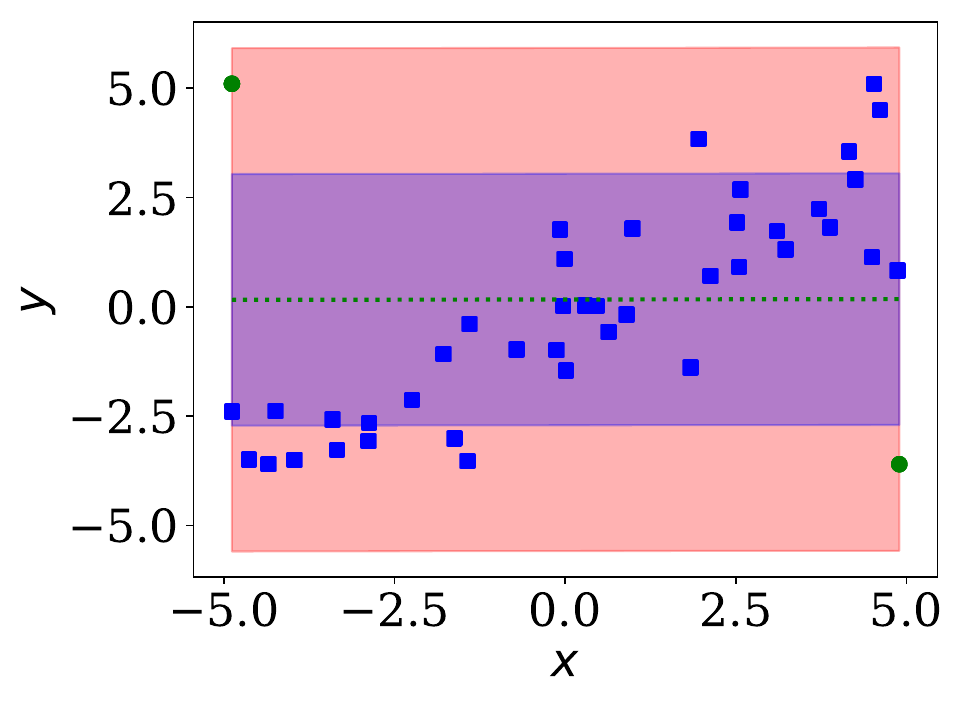}}
\vspace{-.2cm}
\caption{Behavior of BayesClean in a synthetic example. 
In this example, $c_1=1$, $c_2=2$, $\lambda_1=\lambda_2=\beta_1=\beta_2=10^{-6}$, and $T_\text{EM}=300$ (maximum number of iterations when learning the hyperparameters). The blue points represent the clean points, the green points are the poisoning points, the green dotted line is the regression line learned with the complete poisoned training data, the purple region corresponds to the condition $\mathcal{I}_1$, 
and the red region corresponds to $\mathcal{I}_2$. 
(a) $2\%$ of poisoning points. (b) $10\%$ of poisoning points. (c) $18\%$ of poisoning points. It is evident that the predictive variance increases with the number of poisoning points. \vspace{-0.3cm}}
\label{fig:bayesclean}
\end{figure}
}

\vfill

\end{document}